\documentclass{article}

\usepackage{arxiv}

\usepackage{amsmath, amssymb, amsthm}
\usepackage{graphicx, float, booktabs, xcolor, colortbl, multirow, array, tabularx, hyperref, url, geometry, newfloat, caption}
	
\usepackage{xcolor}
\usepackage[ruled]{algorithm2e}
\usepackage[title]{appendix}
\usepackage{subcaption}

\usepackage[english]{babel}

\usepackage{arydshln}

\newcommand{\R}{\mathbb{R}}

\newcommand{\D}{\mathbb{D}}

\newcommand{\Y}{\mathcal{Y}}
\newcommand{\X}{\mathcal{X}}

\newcommand{\E}{\mathcal{E}}

\newcommand{\y}{\mathbf{y}}
\newcommand{\x}{\mathbf{x}}
\newcommand{\z}{\mathbf{z}}
\newcommand{\e}{\mathbf{e}}
\newcommand{\bias}{\mathbf{b}}

\newcommand{\Rec}{\psi}

\newcommand{\addpic}[2][0.49]{\includegraphics[width=#1 \textwidth]{#2}}

\newtheorem{Theorem}{Theorem}[section]

\usepackage{amsthm}

\theoremstyle{definition}
\newtheorem{Definition}{Definition}[section]

\title{Ambiguity in solving imaging inverse problems with deep learning based operators}

\author{
	Davide Evangelista \\
	Department of Mathematics \\
	University of Bologna \\
	\texttt{davide.evangelista5@unibo.it}
	\And
	Elena Morotti \\
	Department of Political and Social Sciences \\
	University of Bologna \\
	\texttt{elena.morotti4@unibo.it}
	\And
	Elena Loli Piccolomini \\
	Department of Computer Science \\
	University of Bologna \\
	\texttt{elena.loli@unibo.it}
	\And
	James Nagy \\
	Department of Mathematics \\
	Emory University \\
	\texttt{jnagy@emory.edu}
}

\begin{document}
\maketitle
\begin{abstract}
In recent years, large convolutional neural networks have been widely used as tools for image deblurring, because of their ability in restoring images very precisely. It is well known that image deblurring is mathematically modeled as an ill-posed inverse problem and its solution is difficult to approximate when noise affects the data. Really, one limitation of neural networks for deblurring is their sensitivity to noise and other perturbations, which can lead to instability and produce poor reconstructions. In addition, networks do not necessarily take into account the numerical formulation of the underlying imaging problem, when trained end-to-end.
In this paper, we propose some strategies to improve stability without losing to much accuracy to deblur images with deep-learning based methods.
First, we suggest a very small neural architecture, which reduces the execution time for training, satisfying a green AI need, and does not extremely amplify noise in the computed image. Second, we introduce a unified framework where a pre-processing step balances the lack of stability of the following, neural network-based, step.
Two different pre-processors are presented: the former implements a strong parameter-free denoiser, and the latter is a variational model-based regularized formulation of the latent imaging problem.
This framework  is also formally characterized by mathematical analysis.  
Numerical experiments are performed to verify the accuracy and stability of the proposed approaches for image deblurring when unknown or not-quantified noise is present; the results confirm that they improve the network stability with respect to noise. In particular, the model-based framework represents the most reliable trade-off between visual precision and  robustness.
\end{abstract}

\keywords{Neural Networks Stability \and Image Deblurring \and Deep Learning \and Inverse Problems in Imaging}

\section{Introduction}\label{sec:introduction}
Image restoration is a discipline within the field of image processing focusing on the removal or reduction of distortions and artifacts from images. This topic is of interest in a wide range of applications, including medical imaging, satellite and aerial imaging, and digital photography.
In this last case,  blurring on images is quite frequent and several factors can cause it. To set some examples, Gaussian blur is caused by the diffraction of light passing through a lens and it is more prevalent in images captured with low-aperture lenses or in situations where the depth of field is shallow, whereas motion blur is due to handheld camera movements or low lighting conditions and slow shutter speeds \cite{hansen_deblurring_images,low_light_deblur_photography_2,deblur_photography}. 
Also noise seriously affects images; it is usually introduced by the acquisition systems.

Researchers have developed a number of algorithms for reducing blur and noise and image restoration is a very active field of research where new methods are continuously being proposed and developed. Such methodologies can be classified into two main categories: model-based and learning-based. The model-based techniques assume that the degradation process is known and it is mathematically described as an inverse problem \cite{bertero2021introduction}. The learning-based methods learn a map between the degraded and clean images during the training phase and use it to deblur new corrupted images \cite{zhang2022deep}.

\subsection*{Model-based mathematical formulation.}
In model-based approaches, denoting by $\X$ the compact and locally connected subset of $\R^n$ of the $\x^{gt}$ ground truth sharp images, the relation between $\x^{gt} \in \X$ and its blurred and noisy observation $\y^\delta$ is formulated as:
\begin{equation}\label{eq:noisy_problem}
    \y^\delta = K \x^{gt} + \e,  \tag{P}
\end{equation}
where $K$ is the known blurring operator and $\e$ represents noise on the image. We can say that, with very high probability, $ ||\e|| \leq \delta$.
In this setting, the goal of model-based image deblurring methods is to compute a sharp and unobstructed image $\x$ given $\y^\delta$ and $K$, by solving the linear inverse problem.
When noise is present, problem \eqref{eq:noisy_problem} is typically reformulated into an optimization problem,
where a data fit measure, namely $\mathcal{F}$, is minimized.
Since the blurring operator $K$ is known to be severely ill-conditioned, a regularization term $\mathcal{R}$ is added to the data-fidelity term $\mathcal{F}$ to avoid noise propagation. The resulting optimization problem is formulated as:
\begin{equation}\label{eq:regularized_problem}
     \x^* = \arg\min_{\x \in \X} \mathcal{F}(K\x, \y^\delta) + \lambda \mathcal{R}(\x),
\end{equation}
where $\lambda > 0$ is the regularization parameter. This optimization problem can be solved using different iterative methods depending on the specific choice for $\mathcal{F}$ and $\mathcal{R}$ \cite{regularization_of_inverse_problems,hansen_deblurring_images,variational_methods_in_imaging}.
We remark that  $\mathcal{F}$ is set as the least-squares function in case of Gaussian noise, whereas te regularization function $\mathcal{R}$ can be tuned by the users according to the imaging properties they desire to enforce. Recently, plug-and-play techniques plug a denoiser, usually a neural network, into an iterative procedure to solve the minimization problem \cite{venkatakrishnan2013plug,kamilov2017plug,cascarano2022plug}. The value of $\lambda$ can also be selected by automatic routines, image-by-image \cite{hansen2010discrete,lazzaro2019nonconvex}. 
These features make model-based approaches mathematically explainable, flexible, and robust. However, a disadvantage is that the final result strongly depends on a set of parameters that are difficult to set up properly.

\subsection*{Deep learning-based formulation.}
In the last decade, deep learning algorithms have been emerging as good alternatives to model-based approaches. 
Disregarding any mathematical blurring operator, convolutional neural networks (NNs) can be trained to identify  patterns characterizing blur on images, thus they can learn several kinds of blur and adapt to each specific imaging task. Large and complex convolutional neural networks, called UNet, have been proposed  to achieve high levels of accuracy, by automatically tuning and defining their inner filters and proper transformations for blur reduction, without needing any parameter setting \cite{nn_deblur2,nn_deblur3,nn_deblur1,nn_deblur4}. 
Indeed, the possibility to process large amounts of data in parallel makes networks highly efficient for image processing tasks and prone to play a key role in the development of new and more advanced techniques in the future. \\

However, challenges and limitations in using neural networks are known in the literature. Firstly, it is difficult to understand and precisely interpret how they are making decisions and predictions, as they act as unexplainable  black boxes  mapping the input image $\y^\delta$ towards $\x^{gt}$ directly.
Secondly, 
neural networks are prone to overfitting, which occurs when they become too specialized for the training samples and perform poorly on new, unseen images. 
Lastly, the high performance of neural networks is typically evaluated only in the so-called \emph{in-domain} case, i.e. the test procedure is performed on images sharing exactly the same corruption with the training samples, hence the impact of unquantified perturbations (as noise can be) has been not widely studied yet.  
In other words, the robustness of NN-based image deblurring with respect to unknown noise is not guaranteed \cite{troublesome_kernel,on_deep_learning_for_inverse_problems,mathematics_of_adversarial_attacks,evangelista2022tobeornottobestable}.


\subsection*{Contributions of the article.}
Motivated by the poor stability but high accuracy of NN-based approaches in solving inverse imaging problems such as deblurring, this paper proposes strategies to improve stability, maintaining good accuracy, acting similarly as regularization functions do in the model-based approach.
Basing on a result showing a trade-off between  stability and accuracy, we  propose to use a very small neural network, in place of the  UNet, which  is less accurate, but it is much more stable than larger networks. Since it has only few parameters to identify, it consumes relatively little time and energy, thus meeting the  green AI principles.

\begin{figure}
    \centering
    \includegraphics[width=0.8\textwidth]{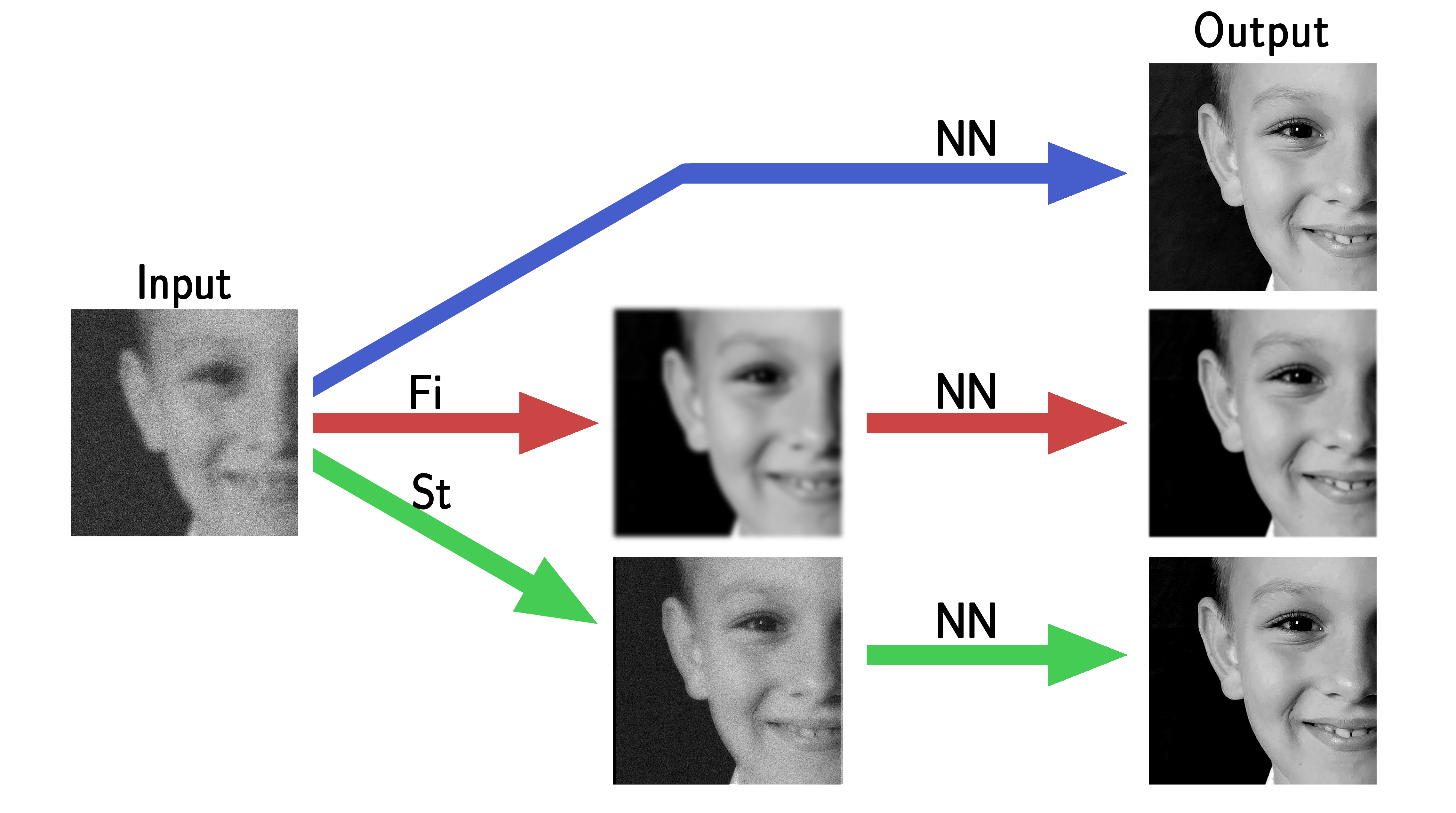}
    \caption{A graphical draft  highlighting the introduction of pre-processing steps Fi and St defining the proposed frameworks FiNN and StNN, respectively.}
    \label{fig:graphAbstract}
\end{figure}

Moreover, we propose two new NN-based schemes, embedding a pre-processing step to face the network instability  when solving deblurring problems as in \eqref{eq:noisy_problem}.
The first scheme, denoted as FiNN, applies a model-free low-pass filter to the datum, before passing it as input to the NN. This is a good approach to be applied whenever an unknown noise is present because it does not need any model information or parameter tuning.
The second scheme, called Stabilized Neural Network (StNN), exploits an estimation of the noise statistics and the mathematical modeling of both noise and  image corruption process.
Figure \ref{fig:graphAbstract} shows a draft of the proposed frameworks. 
whose robustness is evaluated from a theoretical perspective and tested on an image data set.

\subsection*{Structure of the article.}
The work is organized as follows. In Section \ref{sec:reconstructors}, we formulate the NN-based action as an image reconstructor for problem \eqref{eq:noisy_problem}.
In Section \ref{sec:expsetting} we show our experimental set-up and motivate our work on some experiments, thus we state our proposals and derive their main properties in Section \ref{sec:stabilizers}.
Finally, in Section \ref{sec:numerical_experiments} we will report the results of some experiments to test the methods and empirically validate the theoretical analysis, before concluding with final remarks in Section \ref{sec:conclusion}.

\section{Solving imaging inverse problems with Deep Learning based operators}\label{sec:reconstructors}

As stated in  \eqref{eq:noisy_problem}, image restoration is mathematically modeled as an inverse problem which derives 
 from the discretization of Fredholm integral equations, are ill-posed and the noise on the data is amplified in the numerically computed solution of $\y^\delta = K\x^{gt} + \e$. A rigorous theoretical analysis on the solution of such problems with variational techniques which can be formulated as in equation \eqref{eq:regularized_problem} has been performed, both in the continuous and discrete settings, and regularization techniques have been proposed to limit the noise spread in the solution \cite{engl1996regularization,hansen_deblurring_images}. 

At our best knowledge, a similar analysis for deep learning based algorithms is not present in literature and it is quite mysterious how these algorithms behave in presence of noise on the data.  In this paper we use some of the mathematical tools defined and proved in \cite{evangelista2022tobeornottobestable} and we propose here some techniques to limit noise spread. 
More details about the proposed mathematical framework in a more general setting can be found in \cite{evangelista2022tobeornottobestable}.

In the following, if not differently stated, as a vector norm we consider the Euclidean norm.
We first formalize the concept of  reconstructor  associated to \eqref{eq:noisy_problem} with the following definition. 

\begin{Definition}
Denoting by $Rg(K)$ the range of $K$, we call $\Y^\delta = \{ \y^\delta \in \R^n; \inf_{\y \in Rg(K)} || \y - \y^\delta || \leq \delta \}$ the set of corrupted images according to $\delta\ge0$.
Any continuous function  $\Rec: \Y^\delta \to \R^{n}$, mapping $\y^{\delta} = K\x^{gt}+\e $ (where $||\e|| \leq \delta$ with $\delta\geq 0$) to an $\x \in \R^{n}$, is called a reconstructor. 
\end{Definition}
The associated {\it reconstructing error}  is
 \begin{equation}
 \E_\Rec(\x^{gt}, \y^\delta) := ||\Rec(\y^\delta) - \x^{gt} ||.
 \label{eq:rec_error}
\end{equation}
\begin{Definition}
   We quantify the  accuracy of the reconstructor $\Rec$, by defining the measure $\eta > 0$ as:
\begin{equation}\label{eq:accuracy}
    \eta =  \sup_{\x^{gt} \in \X} || \Rec(K\x^{gt}) - \x^{gt} || = \sup_{\x^{gt} \in \X} \E_\Rec(\x^{gt}, \y^0).
\end{equation}
We say that $\Rec$ is $\eta^{-1}$-accurate \cite{engl1996regularization}. 
\end{Definition}

We now consider  a neural network as a particular reconstructor.

\begin{Definition}
Given a neural network architecture $\mathcal{A} = (\nu, S)$ where $\nu = (\nu_0, \nu_1, \dots, \nu_L) \in \mathbb{N}^{L+1}$, $\nu_L=n$, is the width of each layer and $S = (S_{1, 1}, \dots, S_{L, L}), S_{j, k} \in \R^{\nu_j \times \nu_k}$ is the set of matrices representing the skip connections, we define the parametric family $\Xi_\theta^\mathcal{A}$ of neural network reconstructors with architecture $\mathcal{A}$, parameterized by $\theta \in \R^s$, as:
    \begin{equation}
        \Xi_\theta^\mathcal{A} = \{ \Rec_\theta :\Y^\delta \to \R^{n}; \theta \in \R^s \}
    \end{equation}
where $\Rec_\theta(\y^\delta) = \z^L$ is given by:
    \begin{align}
        \begin{cases}
            \z^0 = \y^\delta \\
            \z^{l+1} = \rho(W^l \z^l + \bias^l + \sum_{k=1}^l S_{l, k} \z^k) \quad \forall l = 0, \dots, L-1 
        \end{cases}
    \end{align}
and
 $W^l \in \R^{\nu_{l+1} \times \nu_l}$ is the weight matrix, $\bias^l \in \R^{\nu_{l+1}}$ is the bias vector.    
\end{Definition}


We now analyze the performance of NN-based reconstructors when noise is added to their input.
 
 \begin{Definition}\label{def:const_stab}
    Given $\delta\geq 0$, the $\delta$-stability constant $C^\delta_{\Rec_\theta}$ of an $\eta^{-1}$-accurate reconstructor is defined as:
     \begin{equation}\label{eq:stability_constant}
        C^\delta_{\Rec_\theta} =\sup_{\substack{\x^{gt} \in \X \\ ||\e|| \leq \delta}} \frac{\E_\Rec(\x^{gt}, \y^\delta) - \eta}{|| \e ||_2}.
    \end{equation}
 \end{Definition}  
Since from Definition \ref{def:const_stab} we interestingly observe that the stability constant  amplifies the noise in the data: 

\begin{equation}
    || \Rec_\theta(\y^0 + \e) - \x||_2 \leq \eta + C^\delta_{\Rec_\theta} || \e ||_2 \quad \forall \x \in \X, \ \forall \e \in \R^n, || \e ||_2 \leq \delta,
\end{equation}
with $\y^0$ the  noiseless datum, we can give the following definition:

\begin{Definition}
    Given $\delta\geq 0$, a neural network reconstructor $\Rec_\theta$ is said to be $\delta$-stable if $C^\delta_{\Rec_\theta} \in [0, 1)$.
\end{Definition}

The next theorem states an important relation between the stability constant and the accuracy of a neural network as a solver of an inverse problem .

\begin{Theorem}\label{lemma:stab_vs_acc_tradeoff}
    Let $\Rec_\theta: \R^n \to \R^n$ be an $\eta^{-1}$-accurate reconstructor. Then, for any $x^{gt} \in \X$ and for any $\delta > 0$, $\exists \> \tilde{\e} \in \R^n$ with $|| \tilde{\e} || \leq \delta$ such that
    \begin{equation}
        C^\delta_{\Rec_\theta} \geq \frac{||K^\dagger \tilde{\e}|| - 2\eta}{|| \tilde{\e} ||}
    \end{equation}
    where $K^\dagger$ is the Moore Penrose pseudo-inverse of $K$.
\end{Theorem}
\noindent For the proof see \cite{evangelista2022tobeornottobestable}.\\

We emphasize that, even if  neural networks used as reconstructors do not use any information on the operator $K$, the stability of $\Rec_\theta$ is related to the pseudo-inverse of that operator.

\section{Experimental setting \label{sec:expsetting}}

Here we describe our particular setting using neural networks as  reconstructors for a deblurring application.

\subsection{Newtork architectures}
We have considered three different neural network architectures for deblurring: the widely used UNet \cite{unet}, the recently proposed NAFNet \cite{chen2022simple} and a green AI inspired 3L-SSNet \cite{green_post_processing}. \\

\begin{figure}[b]
    \centering
    \includegraphics[width=\linewidth]{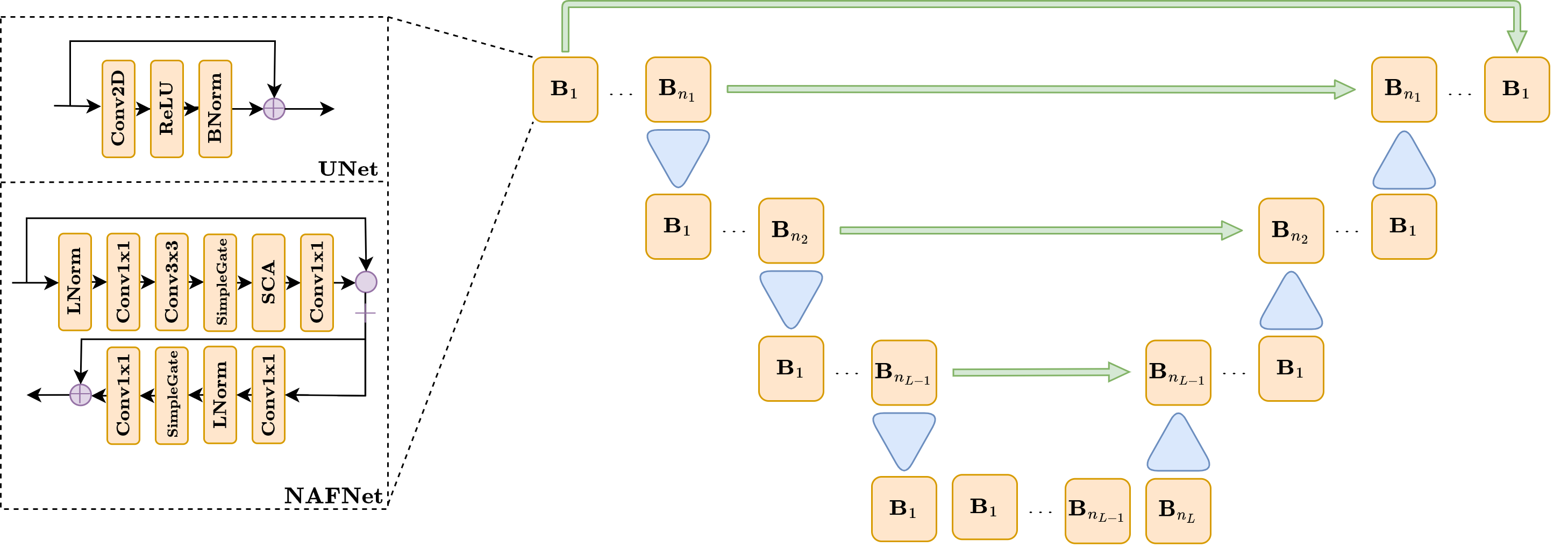}
    \caption{A diagram representing the UNet and NAFNet architectures.}
    \label{fig:UNet_NAFNet_diagram}
\end{figure}

The UNet and NAFNet architectures are complex, multi-scale networks, with similar overall structure but very different behavior. As shown in Figure \ref{fig:UNet_NAFNet_diagram}, both UNet and NAFNet are multi-resolution networks, where the input is sequentially processed by a sequence of blocks $B_1, \dots, B_{n_i}$, $i=1, \dots, L$ and downsampled after that. After $L-1$ downsampling, the image is then sequentially upsampled again to the original shape through a sequence of blocks, symmetrically to what happened in the downsampling phase. At each resolution level $i = 1, \dots, L$, the corresponding image in the downsampling phase is concatenated to the first block in the upsampling phase, to keep the information through the network. Moreover, a skip connection has also been added between the input and the output layer of the model to simplify the training as described in \cite{green_post_processing}. The left-hand side of Figure \ref{fig:UNet_NAFNet_diagram} shows that the difference between UNet and NAFNet is in the structure of each block. In particular, the blocks in UNet are simple Residual Convolutional Layers, defined as a concatenation of Convolutions, ReLU, BatchNormalizations and a skip connection. On the other side, each block in NAFNet is way more complex, containing a long sequence of gates, convolutional and normalization layers. The key propriety of NAFNet, as described in \cite{chen2022simple}, is that no activation function is used in the blocks, since they have been substituted by non-linear gates, thus obtaining improved expressivity and more training efficiency.\\

The 3-layer Single-Scale Network (3L-SSNet) is a very simple model defined, as suggested by its name, by just three convolutional layers, each of them composed by a linear filter, followed by a ReLU activation function and a BatchNormalization layer. Since by construction the network works on single-scale images (the input is never downsampled to low-resolution level, as it is common in image processing), to increase the receptive field of the model the kernel size is crucial. For this reason, we considered a 3L-SSNet with width $[128, 128, 128]$ and kernel size $[9\times 9, 5\times 5, 3\times 3]$, respectively. 

\subsection{Data set} 
\begin{figure}    
\centering
\includegraphics[width = 0.3\textwidth]{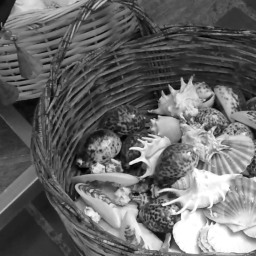} 
\includegraphics[width = 0.3\textwidth]{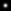} 
\includegraphics[width = 0.3\textwidth]{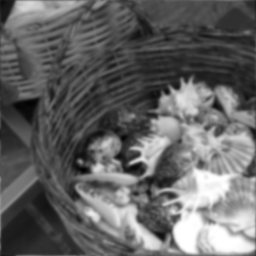} 
\caption{\emph{From left to right:} ground truth clean image, blurring kernel, blurred corrupted image.}
    \label{fig:blur_kernels}
\end{figure}

As a data set for our experiments we choose  the widely-used GoPro \cite{gopro_dataset}, which is composed of a large number of photographic images acquired from a GoPro camera. 
All the images have been cropped  into $256 \times 256$ patches (with no overlapping), converted  into grayscale and normalized into [0,1].  \\ 
We synthesize the blurring of each image according to \eqref{eq:noisy_problem} by considering a Gaussian  corrupting effect, implemented with the $11 \times 11$ Gaussian kernel $\mathcal{G}$ defined as 
\begin{align}\label{eq:gaussian_kernel}
    \mathcal{G}_{i, j} = \begin{cases}
         e^{- \frac{1}{2} \frac{i^2 + j^2}{\sigma_G^2}} \quad & i, j \in \{-5, \dots, 5\}^2 \\
         0 & \text{otherwise}
    \end{cases}
\end{align}
with variance $\sigma_G = 1.3$. The kernel is visualized in Figure \ref{fig:blur_kernels}, together with one of the GoPro images and its blurred counterpart. 

\subsection{Neural networks training and testing \label{sec:experiments}}

To train a Neural Network for deblurring, the set of available images has been split  into train and test subsets, with  $N_{\D} = 2503$ and $N_{\mathbb T} = 1111$ images respectively. Then we consider
 a set $\D = \{ (\y^\delta_i, \x^{gt}_i);\ \x^{gt}_i \in \mathcal{S} \}_{i=1}^{N_\D}$,  for a given $\delta\geq0$. 
Since we set a Mean Squared Error (MSE) loss function, a NN-based reconstructor is uniquely defined  as the solution of:
  \begin{equation}\label{eq:nn_training}
    \min_{\Rec_\theta \in \mathcal{F}_\theta^\mathcal{A}} \sum_{i=1}^{N_\D} || \Rec_\theta(\y^\delta_i) - \x^{gt}_i||_2^2.
\end{equation}
Each network has been trained by performing 50 epochs of Adam optimizer with $\beta_1 = 0.9$, $\beta_2 = 0.9$ and a learning rate of $10^{-3}$. 
We focus on the next two experiments.

{\bf Experiment A}. In this experiment we train  the neural networks on images only corrupted  by blur  ($\delta=0$). To the aim of checking the networks accuracy, defined as in  Section \ref{sec:reconstructors}, we test on no noisy images (\emph{in-domain tests}).
Then, to verify theorem \ref{lemma:stab_vs_acc_tradeoff} we consider test images  with added Gaussian noise, with  $\sigma=0.025$  (\emph{out-of-domain tests}).

{\bf Experiment B}. A common practice for enforcing network stability  is  \emph{noise injection} \cite{training_with_noise_injection}, consisting in training a network by adding noise components to the input. In particular, we have added a vector noise  $\e \sim \mathcal{N}(0, \sigma^2 I)$, with  $\sigma=0.025$. To test the stability of the proposed frameworks with respect to noise, we test with higher noise with respect to training.

\subsection{Robustness of the end-to-end NN approach}

Preliminary results obtained from experiment A are shown in
Figure \ref{fig:experiment_A1}. The first row displays the reconstructions obtained from in-domain tests, where  we can appreciate the accuracy of all the three considered architectures.
In the second row we can see the results obtained from out-of-domain tests, where the noise on the input data strongly corrupts the solution of the ill-posed inverse problem computed by UNet and NAFNet. Confirming what stated by Theorem \ref{lemma:stab_vs_acc_tradeoff}, the best result is obtained with the very light 3L-SSNET, which is the only one able to handle the noise.  

\begin{table}\sffamily
\centering
\begin{tabular}{cccc}
&   UNet  & 3L-SSNet & NAFNet  \\
\rotatebox{90}{In-Domain} &
\addpic[0.27] {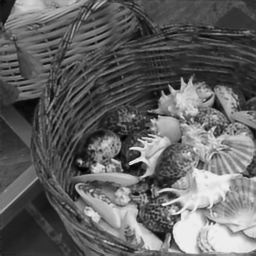} &
\addpic[0.27] {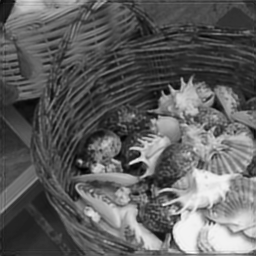} &
\addpic[0.27] {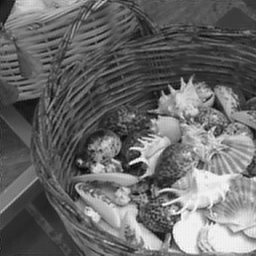} \\
\rotatebox{90}{Out-of-Domain} & 
\addpic[0.27] {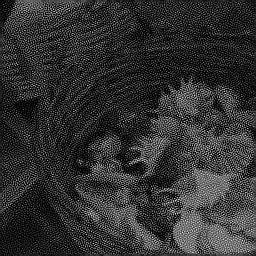} & 
\addpic[0.27] {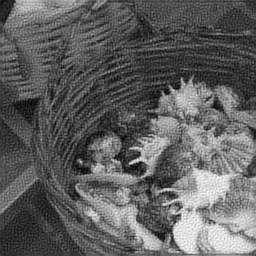}& 
\addpic[0.27] {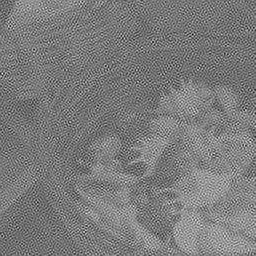}\\
\end{tabular}
\captionof{figure}{Results from  experiment A with the three considered neural networks.
Upper row: reconstruction from no noisy data. Lower row: reconstruction from noisy data ($\delta=0.025$).}
\label{fig:experiment_A1}
\end{table}

\section{Improving noise-robustness in deep learning based reconstructors \label{sec:stabilizers}}

As observed in Section \ref{sec:expsetting}, merely using a neural network to solve an inverse problem is an unstable routine. To enforce the robustness of $\Rec_\theta$ reconstructors, we propose to modify the Deep Learning based approach by introducing a suitable operator, defined in the following as a \emph{stabilizer}, into the reconstruction process.

\begin{Definition}\label{def:stabilizers}
    A continuous functions $\phi: \R^{n} \to \R^{n}$ is called a $\delta$-stabilizer for a  neural network reconstructor $\Rec_\theta: \R^{n} \to \R^{n}$ if
     $\forall \> e \in \R^{n}$ with $||e|| \leq \delta$, $\exists\ L^\delta_\phi \in [0, 1)$ and  $\exists\ e' \in \R^{n}$ with $||e'|| = L^\delta_\phi ||e||$ such that:
        \begin{equation}\label{eq:stabilizer_constant}
            \phi(K \x + \e) = \phi(K\x) + \e'.
        \end{equation}
    In this case, the reconstructor $\bar{\Rec}_\theta = \Rec_\theta \circ \phi$ is said to be $\delta$-stabilized.
    The smallest constant $L^\delta_\phi$  for which the definition holds is the stability constant $C^\delta_\phi$ of  $\phi$.
\end{Definition}


Intuitively, applying a pre-processing $\phi$ with $L^\delta_\phi < 1$ reduces the perturbation of the input data, by converting a noise of amplitude bounded by $\delta$ to a corruption with norm bounded by $\delta L^\delta_\phi$. 
This intuition has been mathematically explained in \cite{evangelista2022tobeornottobestable}, Proposition 4.2, where a relationship between the stability constant of the stabilized reconstructor $\bar{\Rec}_\theta$ and the stability constant of $\Rec_\theta$ has been proved. In particular, if $\bar{\Rec}_\theta = \Rec_\theta \circ \phi$ is a $\delta$-stabilized reconstructor, $L_{\Rec_\theta}^\delta$, $L_\phi^\delta$ are the local Lipschitz constants of $\Rec_\theta$ and $\phi$, respectively, then:
\begin{align}
    C_{\bar{\Rec}_\theta}^\delta \leq L_{\Rec_\theta}^\delta L_\phi^\delta.
\end{align}
As a consequence, if $L^\delta_\phi < 1$, then the stability constant of $\bar{\Rec}_\theta$ is smaller than the Lipschitz constant of $\Rec_\theta$, which implies that $\bar{\Rec}_\theta$ is more stable to input perturbations.

We underline that the $\delta$-stabilizers $\phi$ are effective if they preserve the characteristics and the details of the input image $\y^\delta$. In this paper we focus on the two following  proposals of $\delta$-stabilizers $\phi$.

\subsection{Stabilized Neural Network (StNN) based on the imaging model}

If the blurring operator $K$ is known, it can be exploited to derive a $\delta$-stabilizer function $\phi$. We argue that information on $K$ will contribute to improve the reconstruction accuracy. 
Specifically,  we consider an iterative algorithm, converging to the solution of \eqref{eq:regularized_problem}, represented by the scheme:
\begin{equation}\label{eq:iterative_algorithm}
    \begin{cases}
        \x^{(0)} \in \R^n \\
        \x^{(k+1)} = \mathcal{T}_k(\x^{(k)}; \y^\delta)
    \end{cases}
\end{equation}
where $\mathcal{T}_k$ is  the action of the $k$-th iteration of the algorithm. Given a positive integer $M \in \mathbb{N}$ and a fixed starting iterate $\x^{(0)}$, let us define the $\delta$-stabilizer:
\begin{equation}
    \phi_M(\y^\delta) =\mathop{\bigcirc}\limits_{k=0}^{M-1} \mathcal{T}_k(\x^{(k)}; \y^\delta).
\end{equation}
By definition, $\phi_M$ maps a corrupted image $\y^\delta$ to the solution computed by the iterative solver in $M$ iterations.

Setting as objective function in \eqref{eq:regularized_problem} the Tikhonov-regularized least-squared function:
\begin{align}\label{eq:reg_var_recon}
    \arg\min_{\x \in \R^n} \frac{1}{2} || K\x - \y^\delta ||_2^2 + \lambda ||\x ||_2^2,
\end{align}   
the authors in \cite{evangelista2022tobeornottobestable} showed that it is possible to choose $M$ such that  $L^\delta_{\phi_M} < 1$. Hence, given $\delta$ and $\mathcal{F}_\theta^\mathcal{A}$, it is always possible to use $\phi_M$ as a pre-processing step, stabilizing $\Rec_\theta$. We refer to $\bar \Rec_\theta = \gamma_\theta \circ \phi_M$  as {\em Stabilized Neural Network} (StNN).
In the numerical experiments presented in Section \ref{sec:numerical_experiments}, we use as iterative method for the solution of \eqref{eq:reg_var_recon}  the Conjugate Gradient Least Squares (CGLS) iterative method \cite{hansen2010discrete}.

\subsection{Filtered Neural Network (FiNN)}

The intuition that a pre-processing step should reduce the noise present in the input data naturally leads to our second proposal, implemented by a Gaussian denoising filter. The Gaussian filter is a low-pass filter that reduces the impact of noise on the high frequencies \cite{gonzalez2009digital}. Thus, the resulting pre-processed image is a low-frequency version of $\y^\delta$ and the neural network $\Rec_\theta \in \mathcal{F}_\theta^{\mathcal{A}}$ has to recover the high frequencies corresponding to the image details.
Let $\phi_\mathcal{G}$ represents the operator that applies the Gaussian filter to the input. We will refer to the reconstructor $\bar\Rec_\theta = \Rec_\theta \circ \phi_{\mathcal{G}}$ as {\em Filtered Neural Network} (FiNN).  \\

Note that, even if FiNN is employed to reduce the impact of the noise and consequently to stabilize the network solution, its $L^\delta_\phi$ constant is not smaller than one.
In fact, for any $\e \in \R^n$ with $|| \e || \leq \delta$, it holds:
\begin{align}
    \phi_\mathcal{G}(K\x + \e) = \phi_\mathcal{G}(K\x) + \phi_\mathcal{G}(\e)
\end{align}
as a consequence of the linearity of $\phi_\mathcal{G}$. 


\section{Results}\label{sec:numerical_experiments}

In this Section we present the results obtained in our deblurring experiments described in Section \ref{sec:expsetting}.
To evaluate and compare the deblurred images, we use visual inspection on a selected test image and exploit the Structural Similarity index (SSIM) \cite{SSIM_definition} on the test set.

\subsection*{Results of experiments A \label{sec:expA}} 

We show and comment on the results obtained on experiment A described in Section \ref{sec:experiments}. We remark that aim of these tests is to measure the accuracy of the three considered neural reconstructors and of the stabilizers proposed in Section \ref{sec:stabilizers} and verify their sensitivity to noise in the input data. In a word, how these reconstructors handle the ill-posedness of the imaging inverse problem.

\begin{figure}
\begin{tabular}{ccc}
&  \hspace{3mm} {UNet} & \hspace{4mm} {3L-SSNet} \\
&   In-Domain \hspace{9mm} Out-of-Domain &  In-Domain \hspace{9mm} Out-of-Domain \\
\rotatebox{90}{NN} & 
\includegraphics[trim= 3mm 2mm 5mm 6mm, clip,width=0.22\textwidth]{fig/A1/nn_unet_0_gaussian_recon.png} 
\includegraphics[trim= 3mm 2mm 5mm 6mm, clip,width=0.22\textwidth]{fig/A1/nn_unet_0_g_noise01_gaussian_recon.png} &
\includegraphics[trim= 3mm 2mm 5mm 6mm, clip,width=0.22\textwidth]{fig/A1/nn_ssnet_0_gaussian_recon.png} 
\includegraphics[trim= 3mm 2mm 5mm 6mm, clip,width=0.22\textwidth]{fig/A1/nn_ssnet_0_g_noise01_gaussian_recon.png}\\

\rotatebox{90}{FiNN}& 
\includegraphics[trim= 3mm 2mm 5mm 6mm, clip,width=0.22\textwidth]{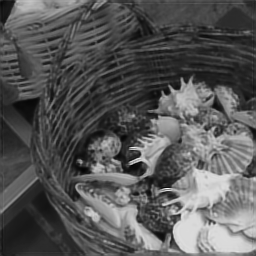} 
\includegraphics[trim= 3mm 2mm 5mm 6mm, clip,width=0.22\textwidth]{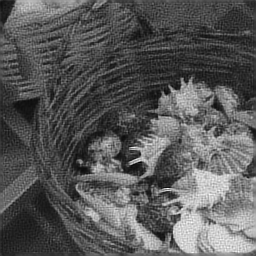} &
\includegraphics[trim= 3mm 2mm 5mm 6mm, clip,width=0.22\textwidth]{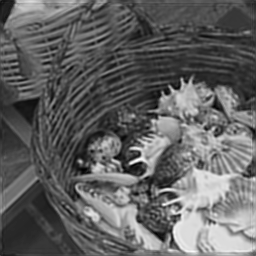} 
\includegraphics[trim= 3mm 2mm 5mm 6mm, clip,width=0.22\textwidth]{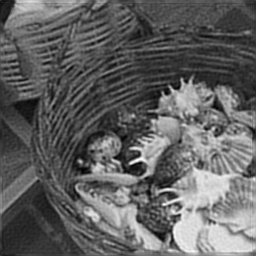}\\

\rotatebox{90}{StNN}& 
\includegraphics[trim= 3mm 2mm 5mm 6mm, clip,width=0.22\textwidth]{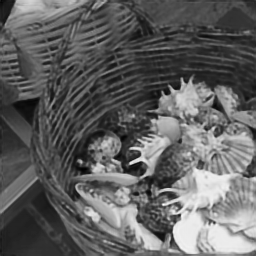} 
\includegraphics[trim= 3mm 2mm 5mm 6mm, clip,width=0.22\textwidth]{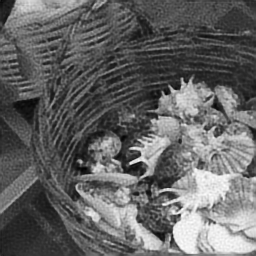} &
\includegraphics[trim= 3mm 2mm 5mm 6mm, clip,width=0.22\textwidth]{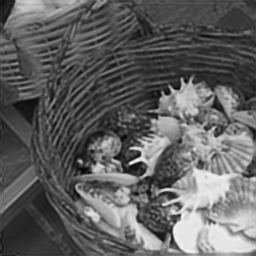} 
\includegraphics[trim= 3mm 2mm 5mm 6mm, clip,width=0.22\textwidth]{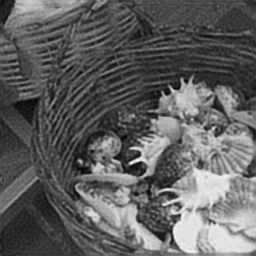}\\
\end{tabular}
\caption{Results from experiment A with UNet and 3L-SSNet.}
\label{fig:experiment_A2}
\end{figure}

To this purpose, we visually compare the reconstructions of a single test image by the UNet and 3L-SSNet in Figure \ref{fig:experiment_A2}. The first row (which replicates some of the images of Figure \ref{fig:experiment_A1})
shows the results of the deep learning based reconstructors, where the out-of-domain images are clearly damaged by the noise. The FiNN and, particularly, the StNN stabilizer drastically reduce noise, producing accurate results even for out-of-domain tests.

In order to analyze the accuracy and stability of our proposals, we compute  the empirical accuracy $\hat{\eta}^{-1}$ and the empirical stability constant $\hat{C}^\delta_\Rec$, respectively defined as:
\begin{equation}\label{eq:empirical_accuracy}
    \hat{\eta}^{-1} = \Bigl( \> \sup_{\x \in \mathcal{S}_\mathbb{T}} || \Rec(K\x) - \x ||_2 \Bigr)^{-1}
\end{equation}
and
\begin{equation}\label{eq:empirical_stability}
    \hat{C}^\delta_\Rec = \sup_{\x \in \mathcal{S}_\mathbb{T}} \frac{|| \Rec(K\x + \e) - \x ||_2 - \hat{\eta}}{|| \e ||_2}
\end{equation}
where $\mathcal{S}_\mathbb{T} \subseteq \X$ is the test set and $\e$ is a noise realization from $\mathcal{N}(0, \sigma^2 I)$ with $||e||_2\leq \delta$ (different for any datum $x \in \mathcal{S}_\mathbb{T}$). 

The computed values are reported in Table \ref{tab:stability_constants}.
Focusing on the estimated accuracies, the results confirm that NN is the most accurate method, followed by NAFNet and 3L-SSNet, as expected. As a consequence of Theorem \ref{lemma:stab_vs_acc_tradeoff}, the values of the stability constant $\hat{C}^\delta_\Rec$ are in reverse order: the most accurate is the less stable (notice the very high value  of $\hat{C}^\delta_\Rec$ for NN!). By applying the stabilizers, the accuracy is slightly lower but the stability is highly improved  (in  most of the cases the constant is less than one), confirming the efficacy of the proposed solutions to handle noise and, at the same time, maintain good image quality. 
In particular, StNN is a stable reconstructor independently from the architecture.

\begin{table}[hbt]
    \centering
    \begin{tabular}{l ccc ccc}
\toprule
 \arrayrulecolor{black!30}
  &   \multicolumn{3}{c}{ $\hat{\eta}^{-1}$}  & \multicolumn{3}{c}{ $\hat{C}^\delta_\Rec$ } \\
\cmidrule(lr){2-4}   \cmidrule(lr){5-7}   
  &   $NN$ & $FiNN$ & $StNN$ & $NN$ & $FiNN$ & $StNN$ \\
\arrayrulecolor{black} \bottomrule    
UNet &   0.118 & 0.085 & 0.087 & 36.572 & 2.519 & 0.878 \\
3L-SSNet &   0.082 & 0.055 & 0.072 & 2.563 & 0.148 & 0.243 \\
NAFNet &  0.104 & 0.080 & 0.078 & 15.624 & 1.053 & 0.434 \\
\bottomrule
    \end{tabular}
    \caption{Estimated accuracy and stability constants for experiment A on out-of-domain test (input images corrupted by noise with $\delta=2.56$). }
    \label{tab:stability_constants}
\end{table}

\begin{figure}
\centering
\begin{tabular}{cccc}
&   NN  &  FiNN & StNN\\

\rotatebox{90}{\qquad UNet} & 
\includegraphics[width=0.28\textwidth]{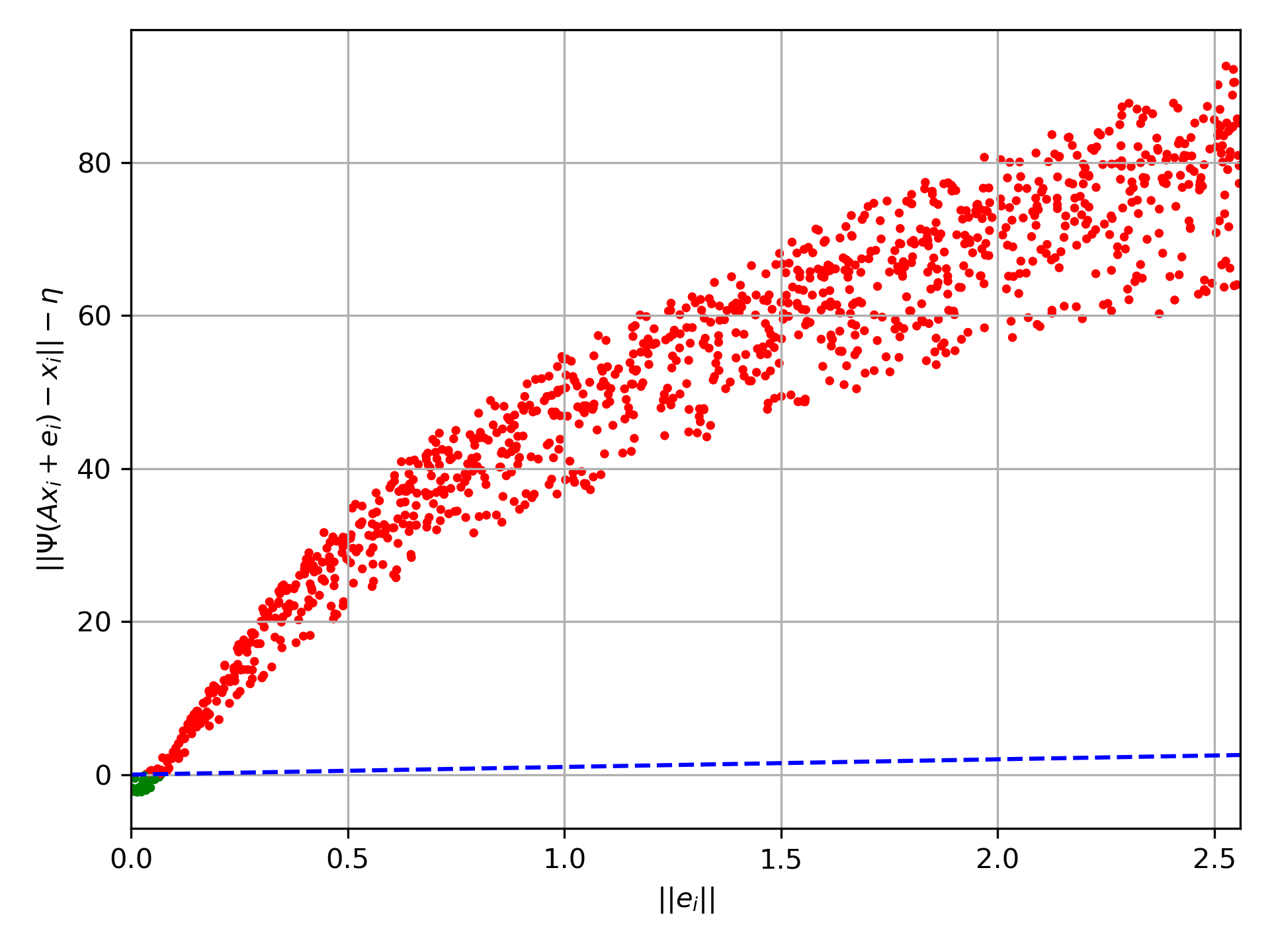} &
\includegraphics[width=0.28\textwidth]{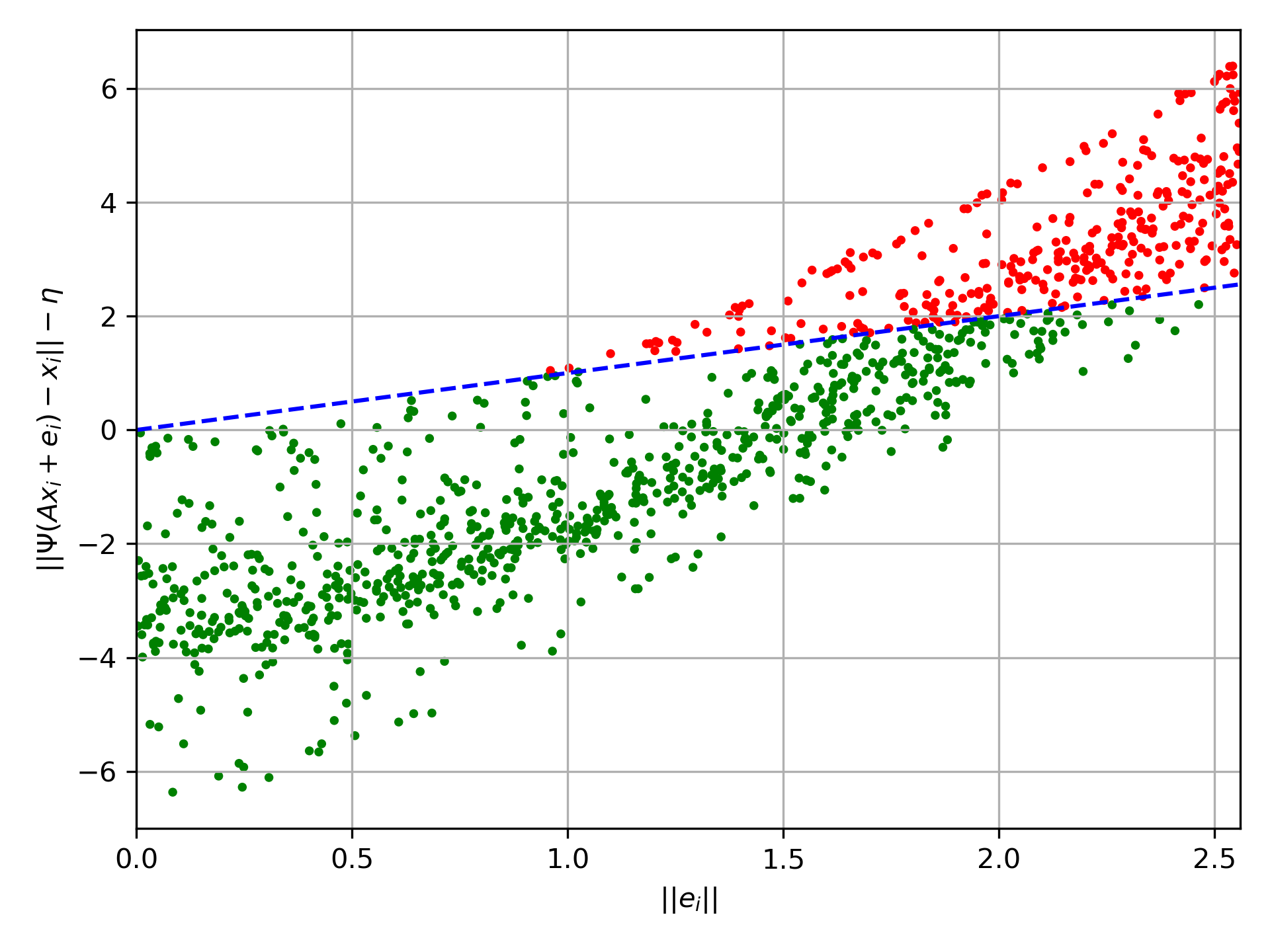} &
\includegraphics[width=0.28\textwidth]{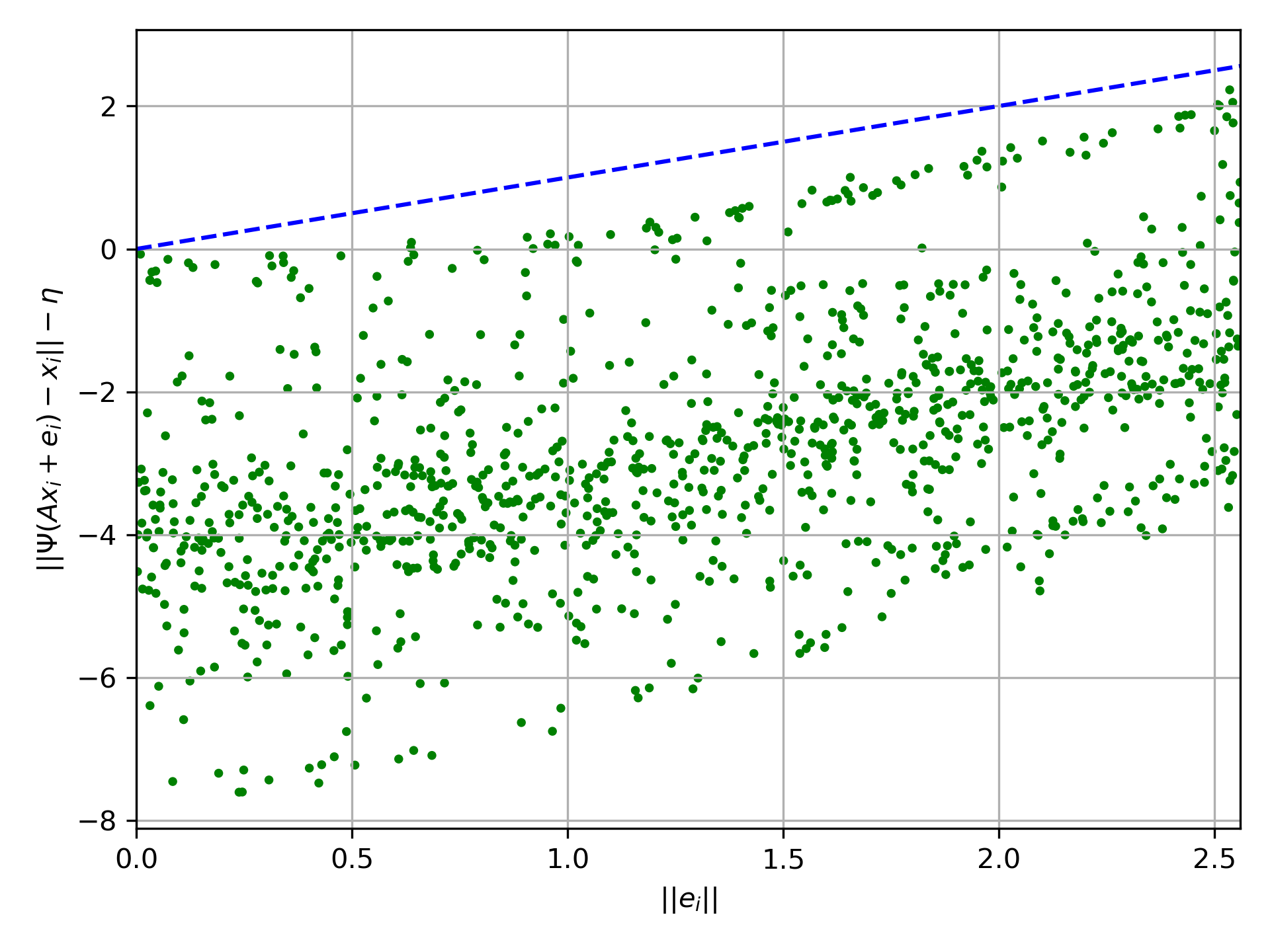} \\

\rotatebox{90}{\qquad 3L-SSNet} & 
\includegraphics[width=0.28\textwidth]{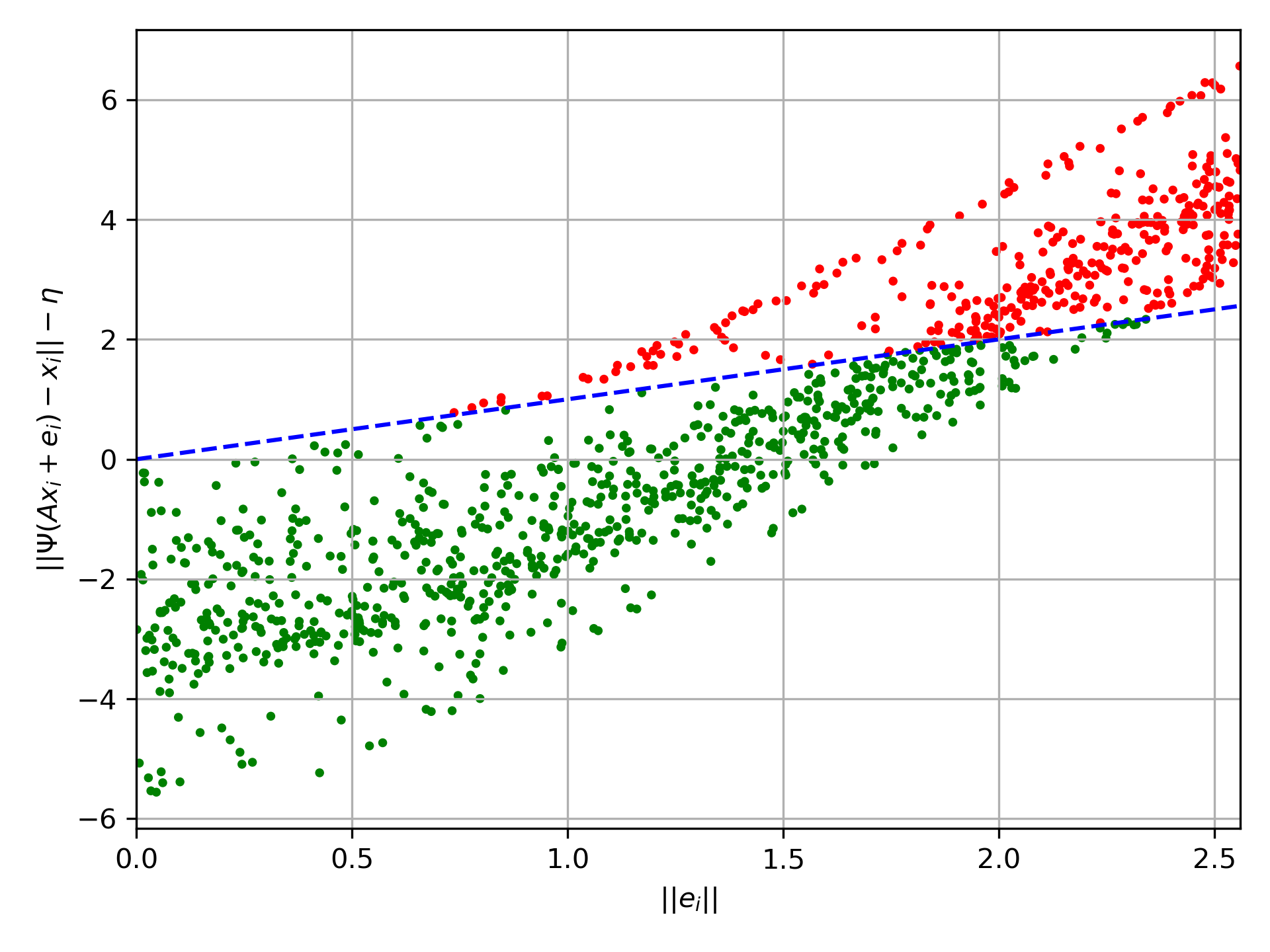} &
\includegraphics[width=0.28\textwidth]{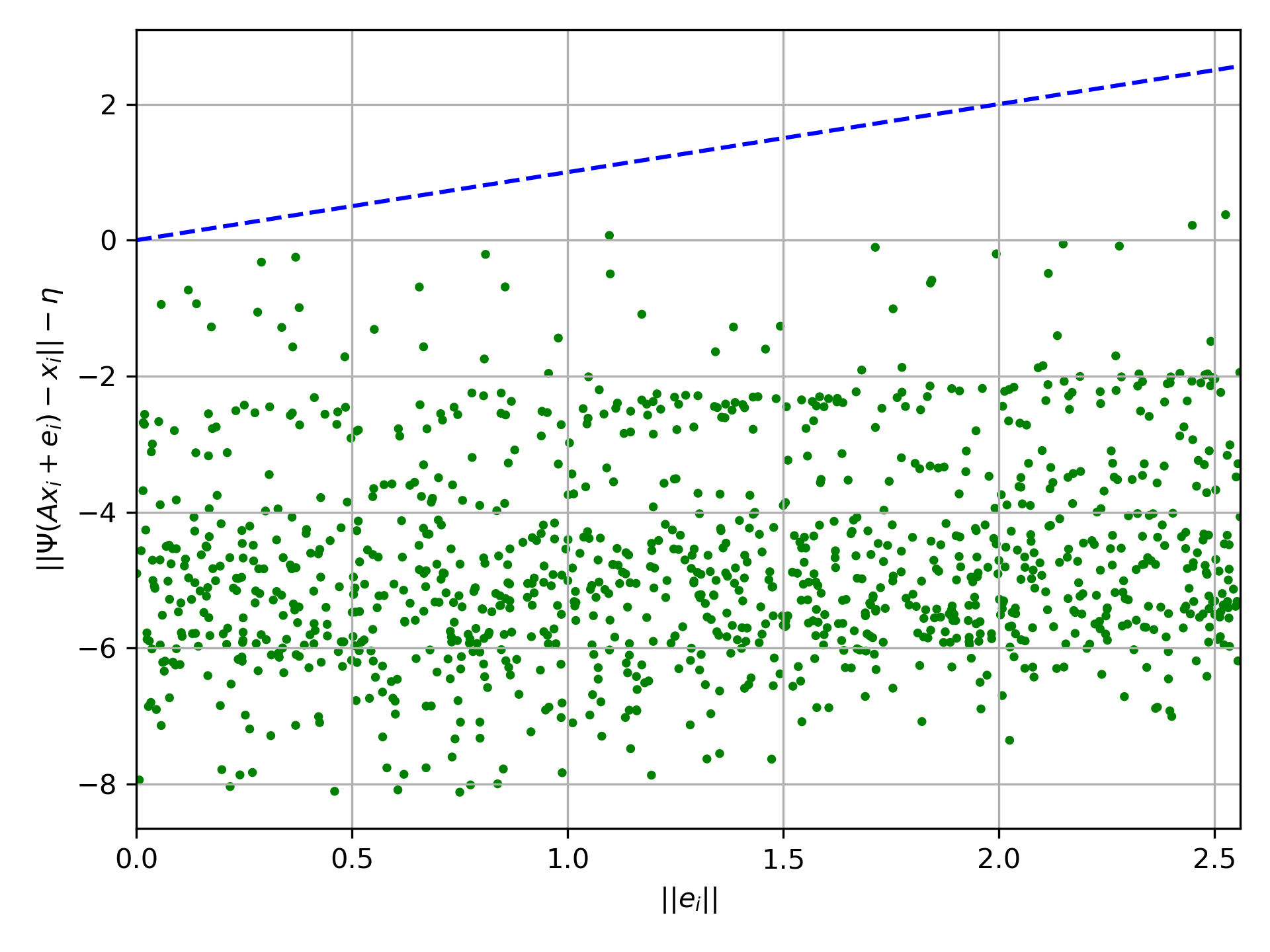} &
\includegraphics[width=0.28\textwidth]{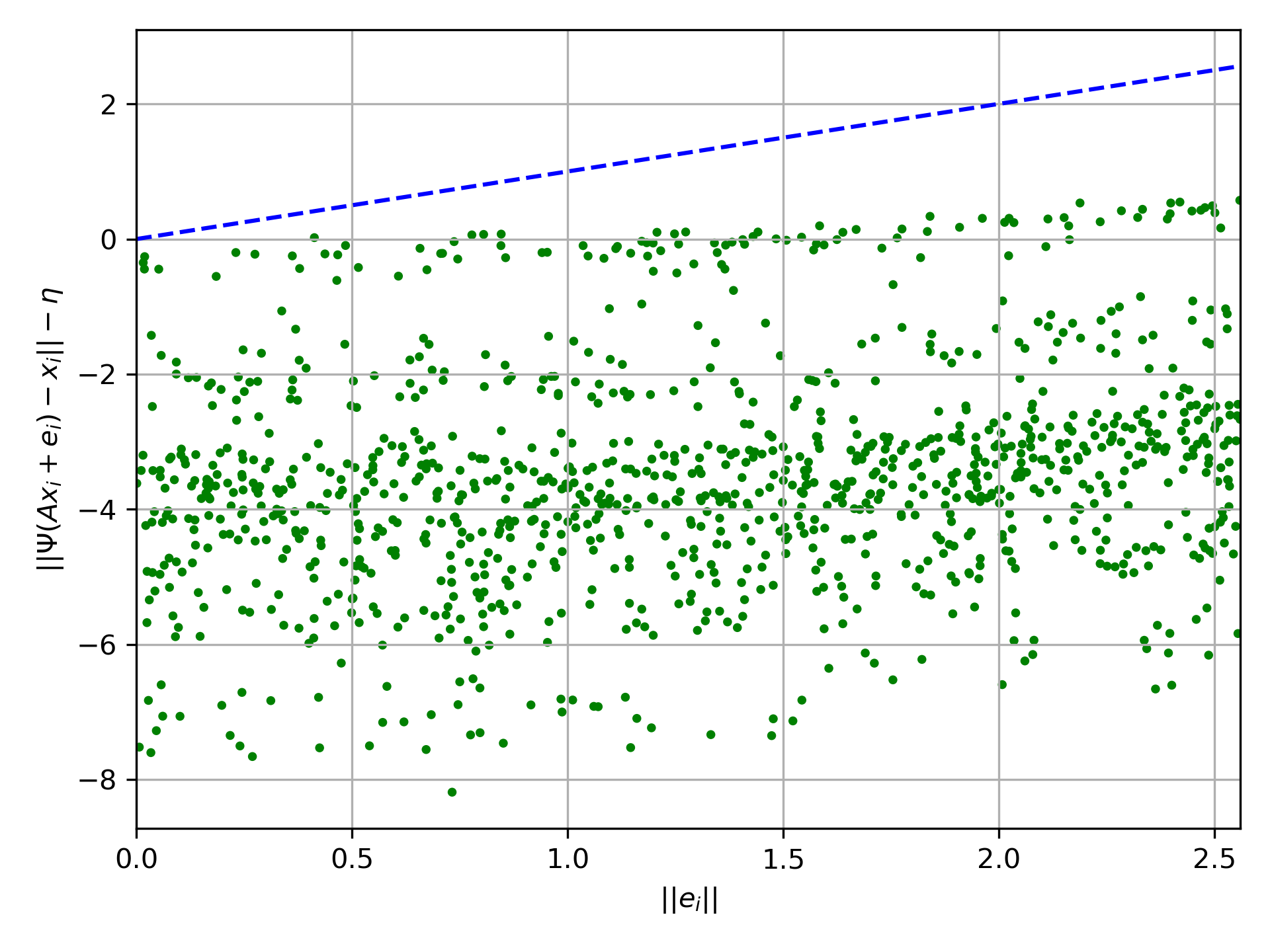} \\
\end{tabular}
\caption{Results from experiment A. Plot of $\E_\Rec(\x^{gt}, \y^\delta) - \eta$  vs.   $\|e\|$ for all the test images. The blue dashed line represents the bisect.}
\label{fig:experiment_A3}
\end{figure}

To analyse the stability of the test set with respect to noise, we have plotted in Figure \ref{fig:experiment_A3}, for each test image, $\E_\Rec(\x^{gt}, \y^\delta) - \hat{\eta}$  vs.   $\|e\|$, where the reconstruction error is defined in \eqref{eq:rec_error}. With green and red dots we have plotted the experiments with stability constant less and greater than one, respectively and with the  blue dashed line the bisect. We notice that the values reported in Table \ref{tab:stability_constants} for the empirical stability constant computed as supremum (see  Equation \eqref{eq:empirical_stability}) are not outliers but they are representative of the results of the whole test set.

\subsection{Results of experiment B \label{sec:expB}}

In this experiment we used noise injection in the neural networks training, as described in Section \ref{sec:experiments}. This quite common strategy reduces the networks accuracy but improve their stability with respect to noise. However, we show that the reconstructions are not totally satisfactory when we test on out-of-domain images, i.e. when input images are affected by  noise of different intensities with respect to training.

\begin{figure}
\centering
\begin{tabular}{c cc}
&    In-Domain & Out-of-Domain \\
& NN \hspace{12mm} FiNN \hspace{10mm} StNN &  NN \hspace{12mm}  FiNN \hspace{10mm} StNN \\

\rotatebox{90}{UNet} &
\includegraphics[trim= 3mm 2mm 5mm 6mm, clip,width=0.13\textwidth]{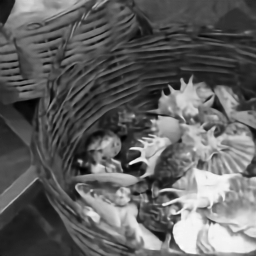}  
\includegraphics[trim= 3mm 2mm 5mm 6mm, clip,width=0.13\textwidth]{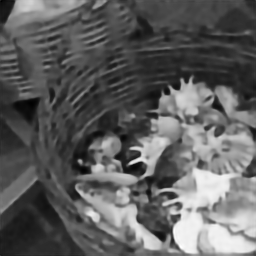} 
\includegraphics[trim= 3mm 2mm 5mm 6mm, clip,width=0.13\textwidth]{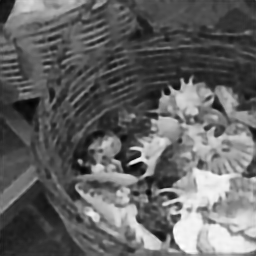} 
&
\includegraphics[trim= 3mm 2mm 5mm 6mm, clip,width=0.13\textwidth]{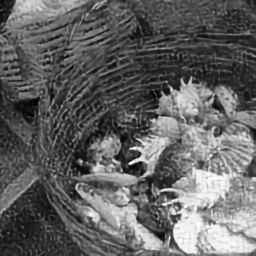}
\includegraphics[trim= 3mm 2mm 5mm 6mm, clip,width=0.13\textwidth]{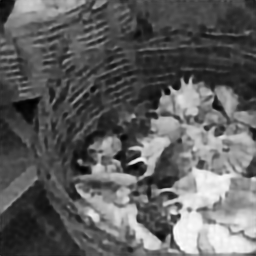} 
\includegraphics[trim= 3mm 2mm 5mm 6mm, clip,width=0.13\textwidth]{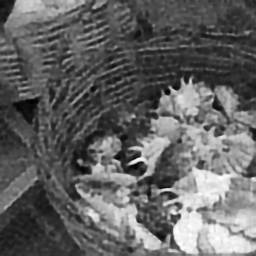}\\

\rotatebox{90}{3L-SSNet} &
\includegraphics[trim= 3mm 2mm 5mm 6mm, clip,width=0.13\textwidth]{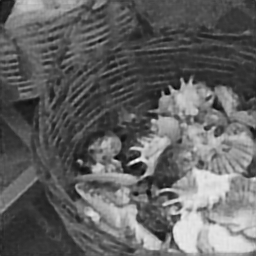}  
\includegraphics[trim= 3mm 2mm 5mm 6mm, clip,width=0.13\textwidth]{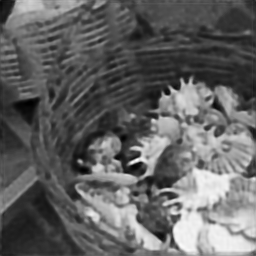} 
\includegraphics[trim= 3mm 2mm 5mm 6mm, clip,width=0.13\textwidth]{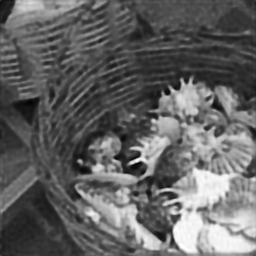} 
&
\includegraphics[trim= 3mm 2mm 5mm 6mm, clip,width=0.13\textwidth]{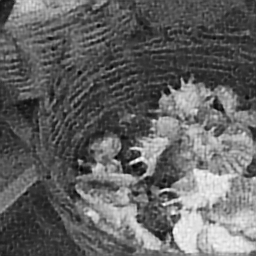}
\includegraphics[trim= 3mm 2mm 5mm 6mm, clip,width=0.13\textwidth]{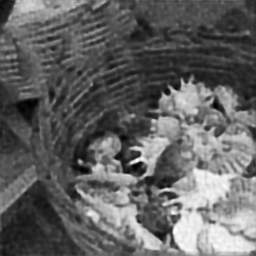} 
\includegraphics[trim= 3mm 2mm 5mm 6mm, clip,width=0.13\textwidth]{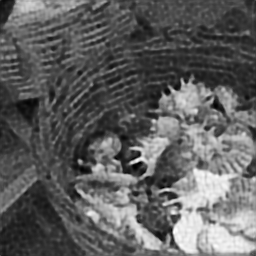}\\

\rotatebox{90}{NAFNet}  &
\includegraphics[trim= 3mm 2mm 5mm 6mm, clip,width=0.13\textwidth]{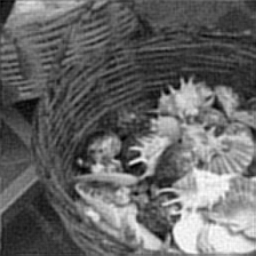}  
\includegraphics[trim= 3mm 2mm 5mm 6mm, clip,width=0.13\textwidth]{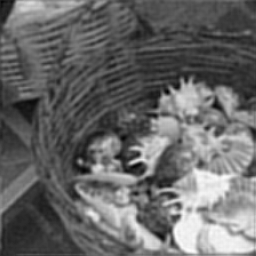} 
\includegraphics[trim= 3mm 2mm 5mm 6mm, clip,width=0.13\textwidth]{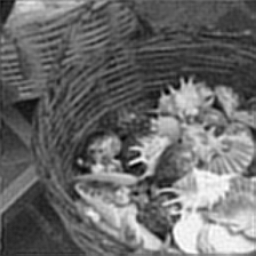} 
&
\includegraphics[trim= 3mm 2mm 5mm 6mm, clip,width=0.13\textwidth]{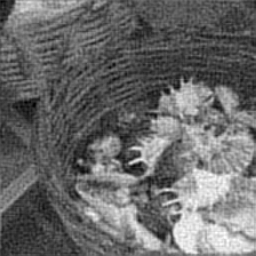}
\includegraphics[trim= 3mm 2mm 5mm 6mm, clip,width=0.13\textwidth]{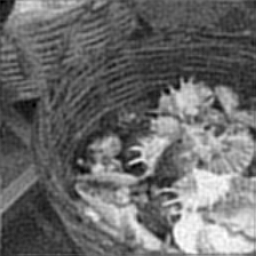}  
\includegraphics[trim= 3mm 2mm 5mm 6mm, clip,width=0.13\textwidth]{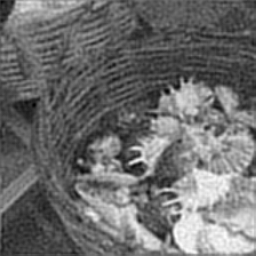}\\
\end{tabular}
\caption{Results from the experiment B. On the left, tests with images with the same noise as in the training ($\delta=0.025$). On the right, tests on images with higher noise ($\delta=0.075$).}
\label{fig:experiment_B1}
\end{figure}

Figure \ref{fig:experiment_B1} displays the reconstructions obtained by testing with both in-domain (on the left) and out-of-domain (on the right) images. Even if the NN reconstructions  (column 4) are not so injured by noise as in experiment A (see Figure \ref{fig:experiment_A1}), however noise artifacts are clearly visible, especially in UNet and NAFNet. Both the stabilizers proposed act efficiently and remove most of the noise. We observe that the restorations obtained with FiNN are smoother but also more blurred with respect to the ones computed by StNN.

\begin{figure}
\centering
\begin{tabular}{c cc}
& In-Domain & Out-of-Domain \\

\rotatebox{90}{\hspace{5mm}UNet} &
\includegraphics[width=0.45\textwidth]{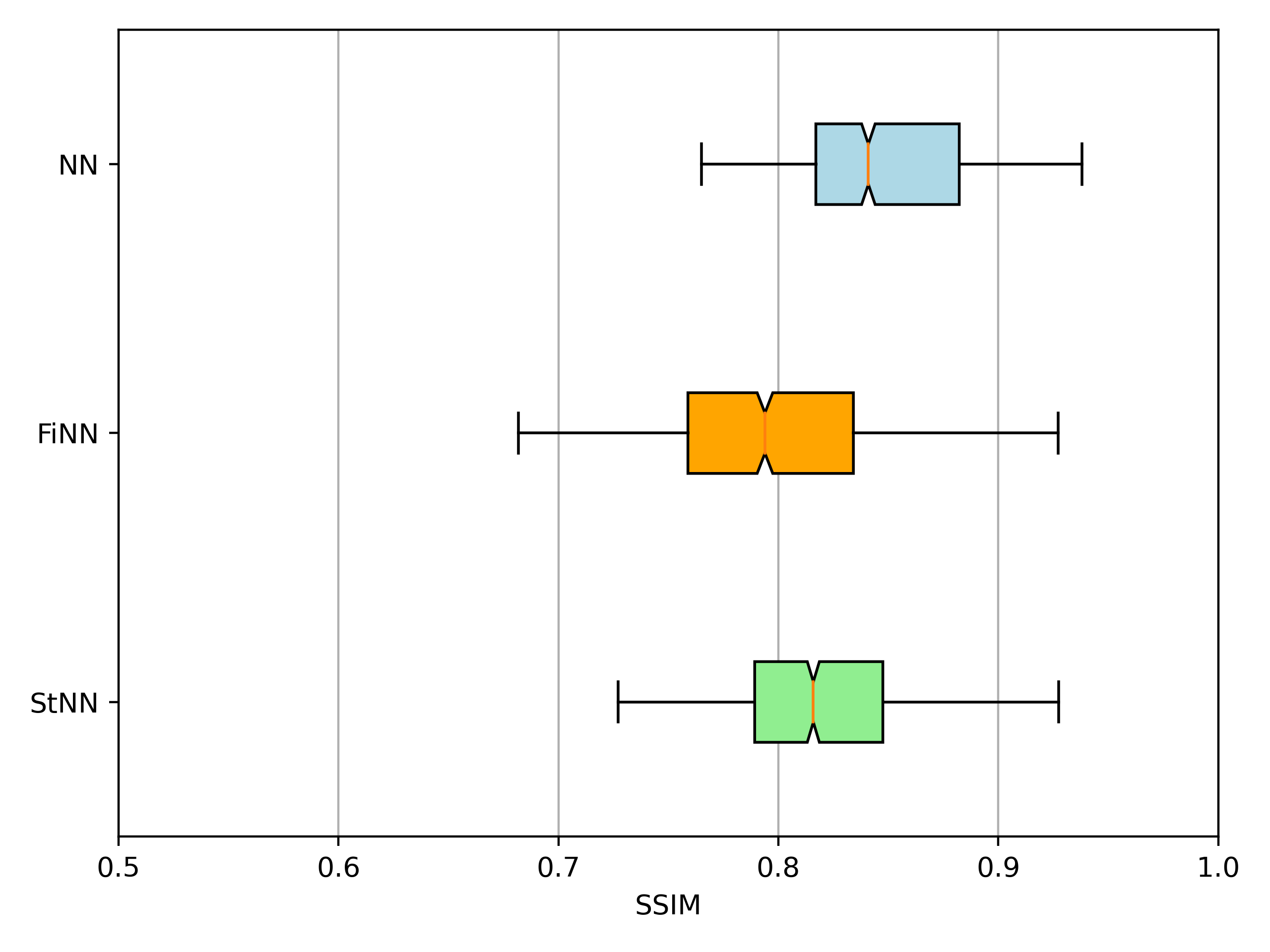} &
\includegraphics[width=0.45\textwidth]{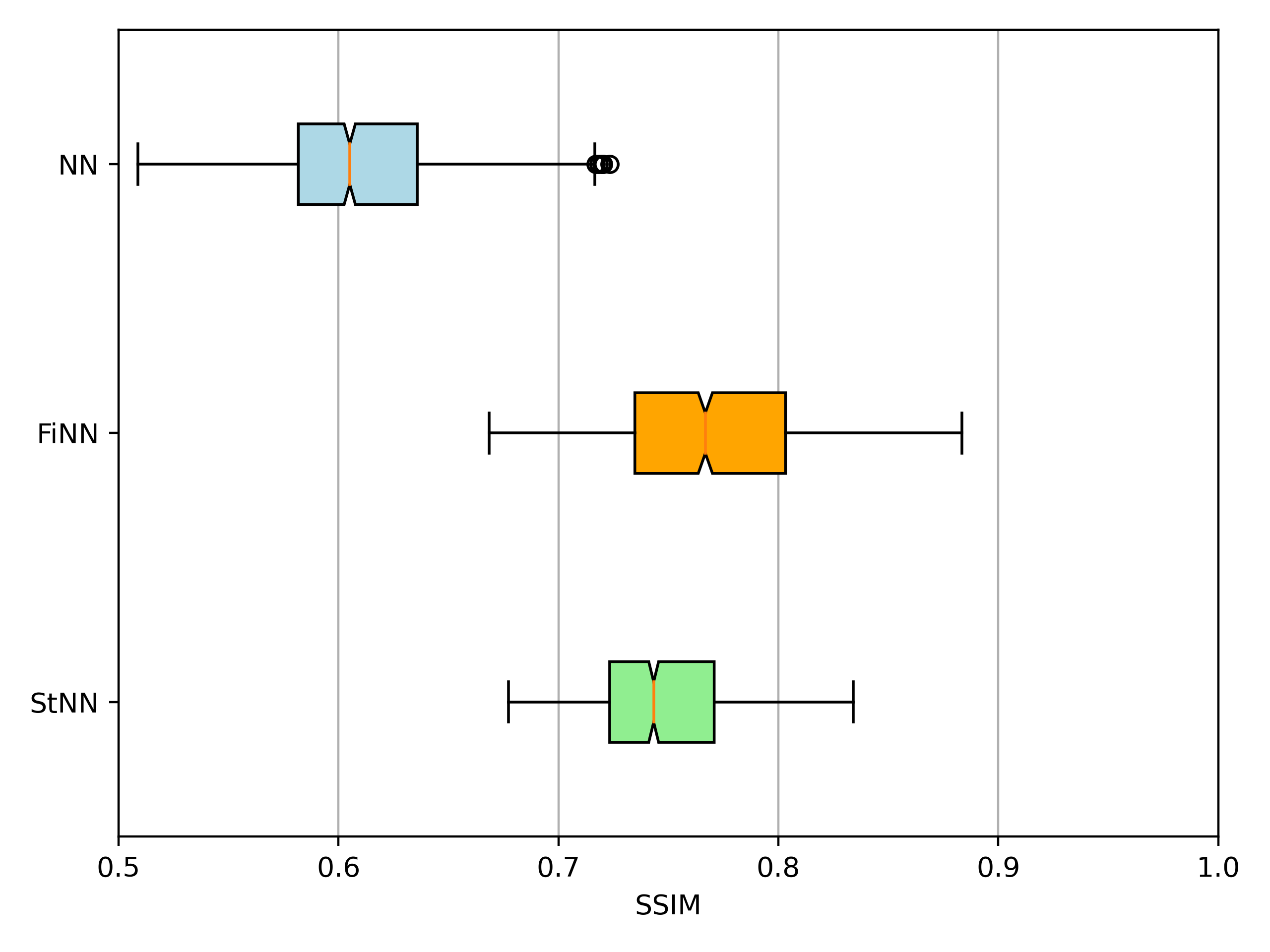}\\

\rotatebox{90}{\hspace{5mm}3L-SSNet} &
\includegraphics[width=0.45\textwidth]{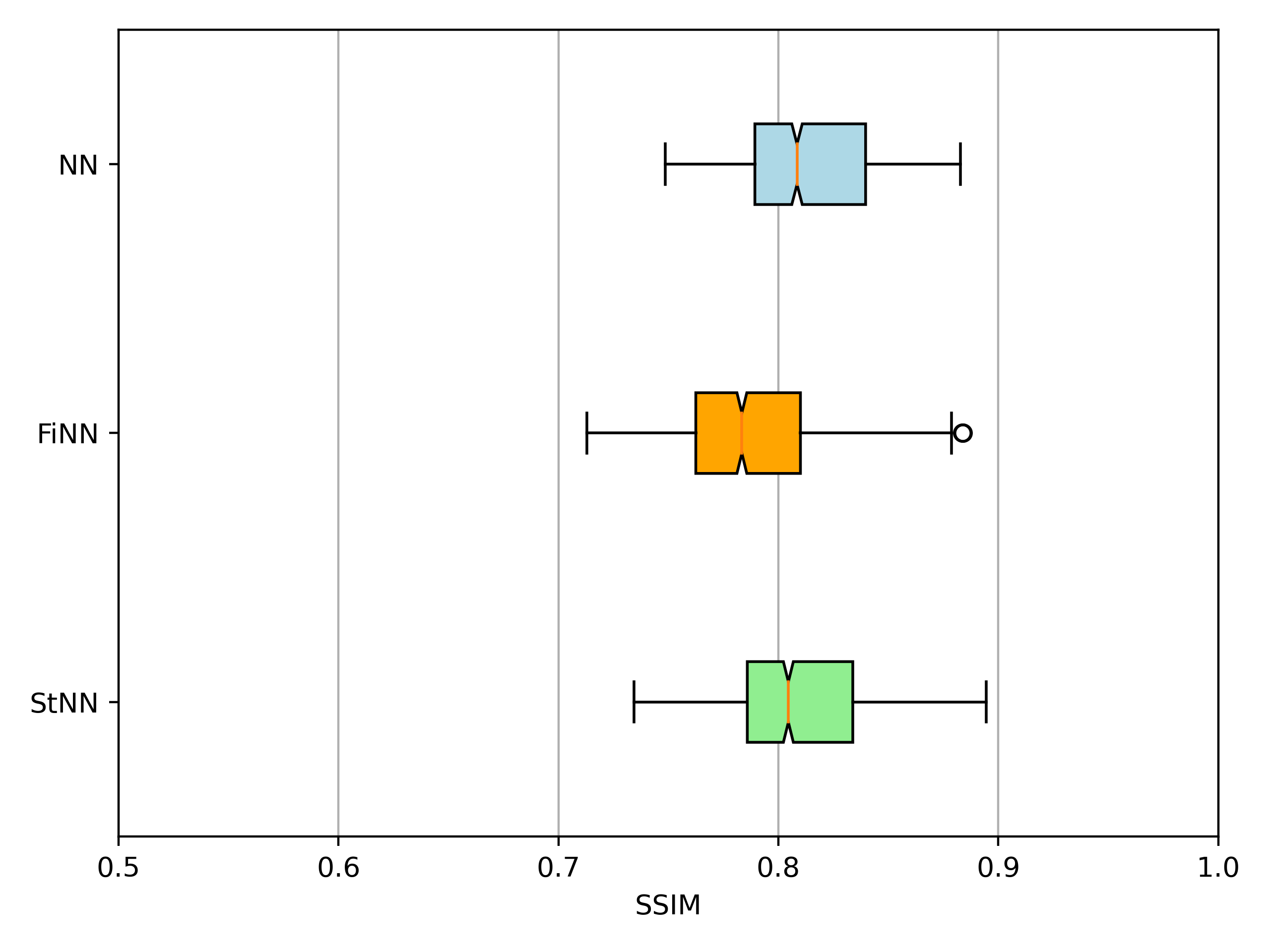} &
\includegraphics[width=0.45\textwidth]{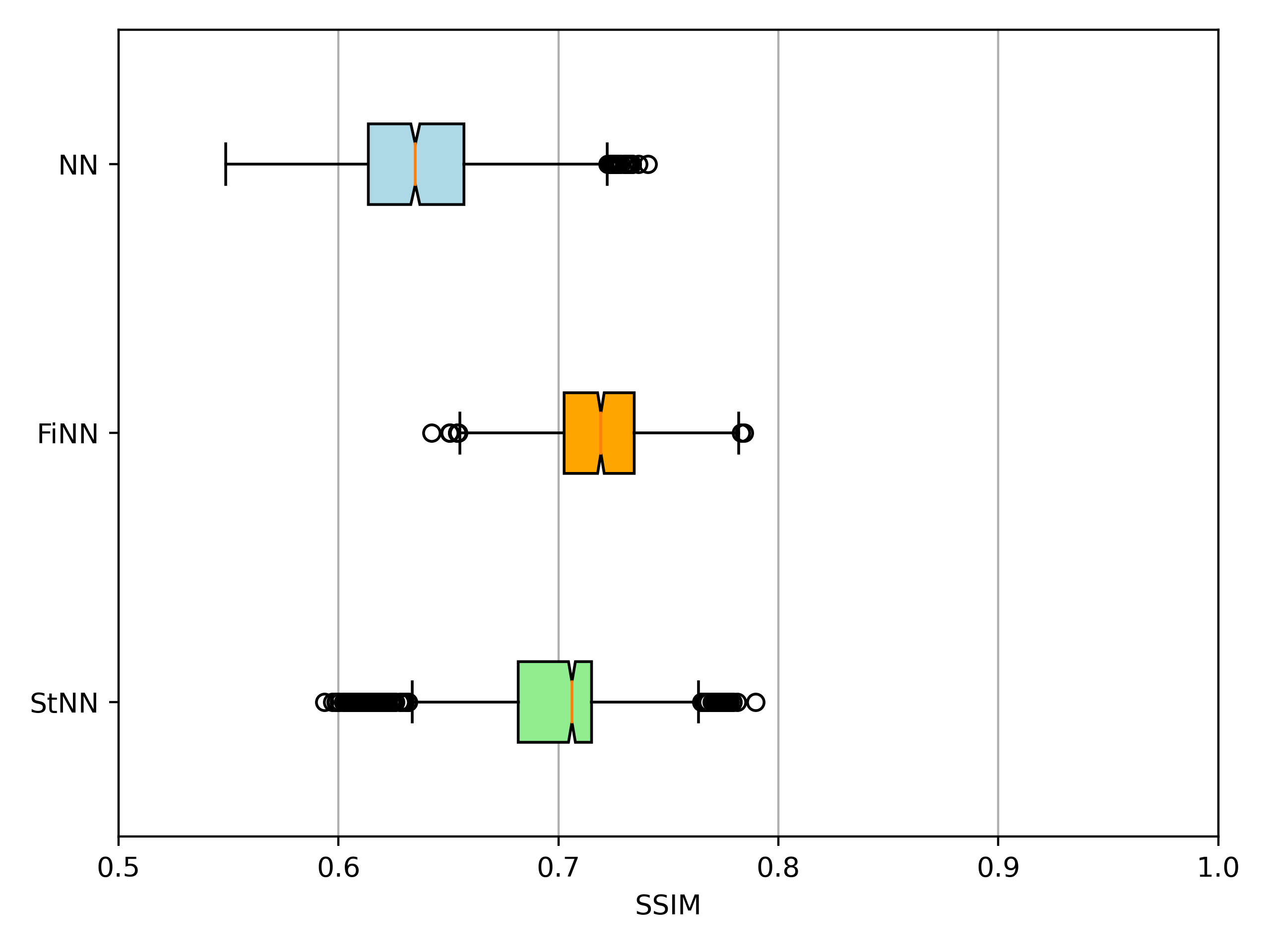}\\

\rotatebox{90}{\hspace{5mm}NAFNet}  &
\includegraphics[width=0.45\textwidth]{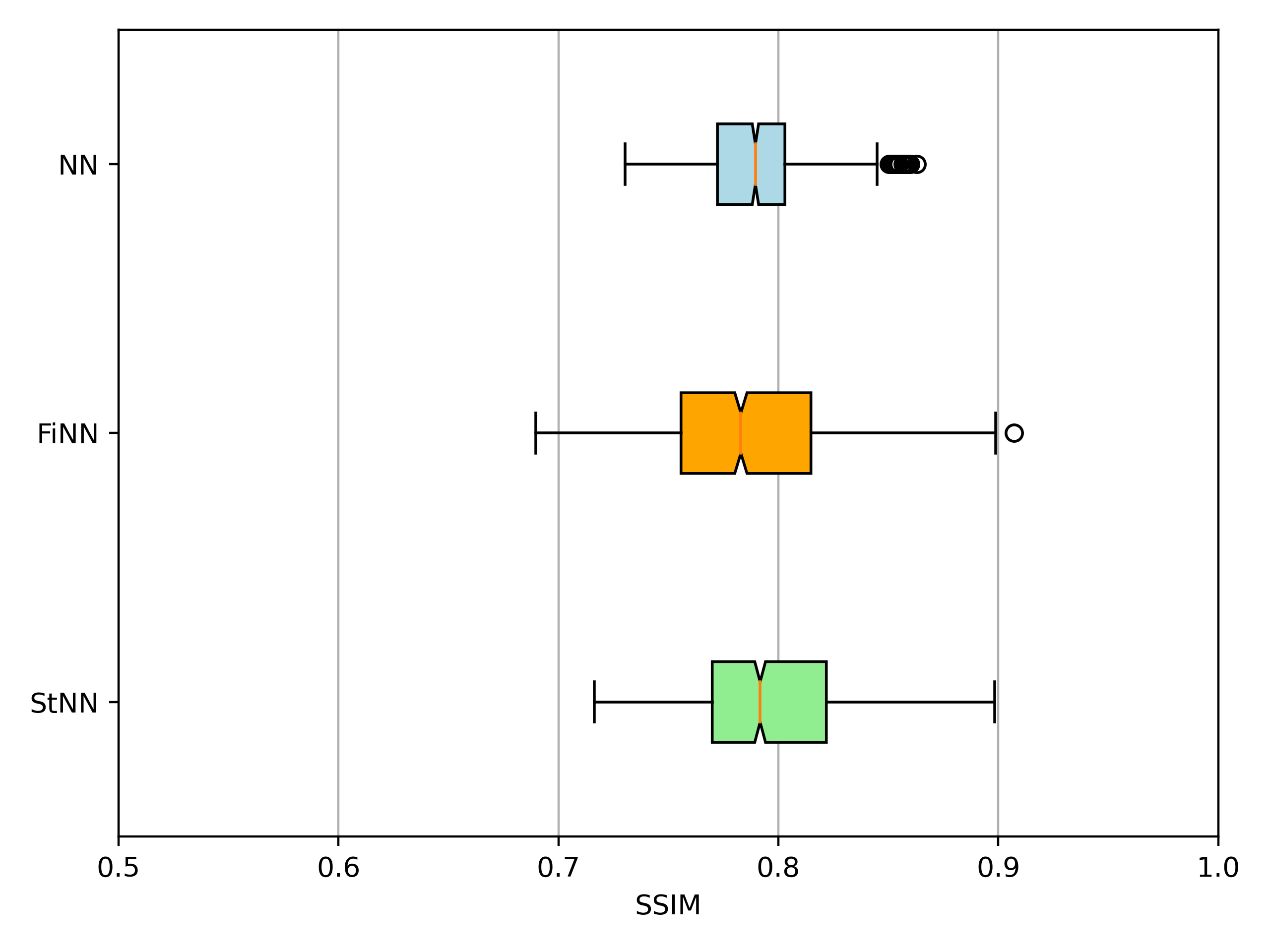} &
\includegraphics[width=0.45\textwidth]{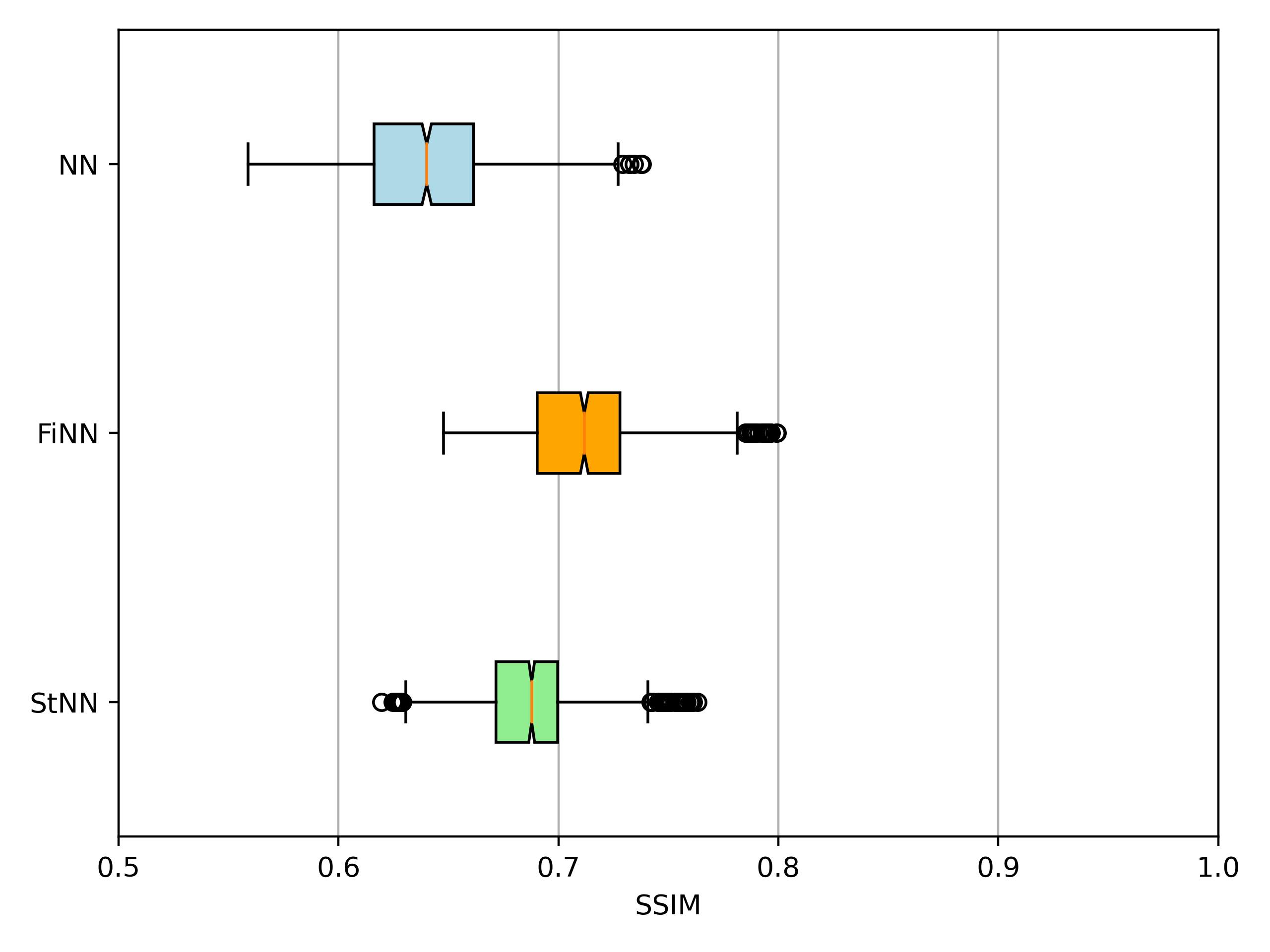}\\
\end{tabular}
\caption{Boxplots for the SSIM values in experiment B. The light blue, orange and green boxplots represent the results computed by NN, FiNN and StNN, respectively.}
\label{fig:experiment_B2}
\end{figure}

An overview of the tests is displayed by the boxplots of the SSIM values sketched in Figure \ref{fig:experiment_B2}.
The light blue, orange and green boxes represent the results obtained with NN, FiNN and StNN methods, respectively.
They confirm that the neural networks performance worsens with noisy data (see the different positions of light blue boxes from the left to the right column), whereas the proposed  frameworks including FiNN and StNN  are far more stable.

\subsection{Analysis with noise varying on the test set \label{ssec:results3}}

Finally, we have analysed the performance of the methods when the input image $\y^\delta$ is corrupted by noise $\| \e \| $  from $\mathcal{N}(0, \sigma^2 I)$, with $\sigma$ varying.

\begin{figure}
\centering
\begin{tabular}{c c c}
UNet & 3L-SSNet   & NAFNet  \\
\includegraphics[width=0.3\textwidth]{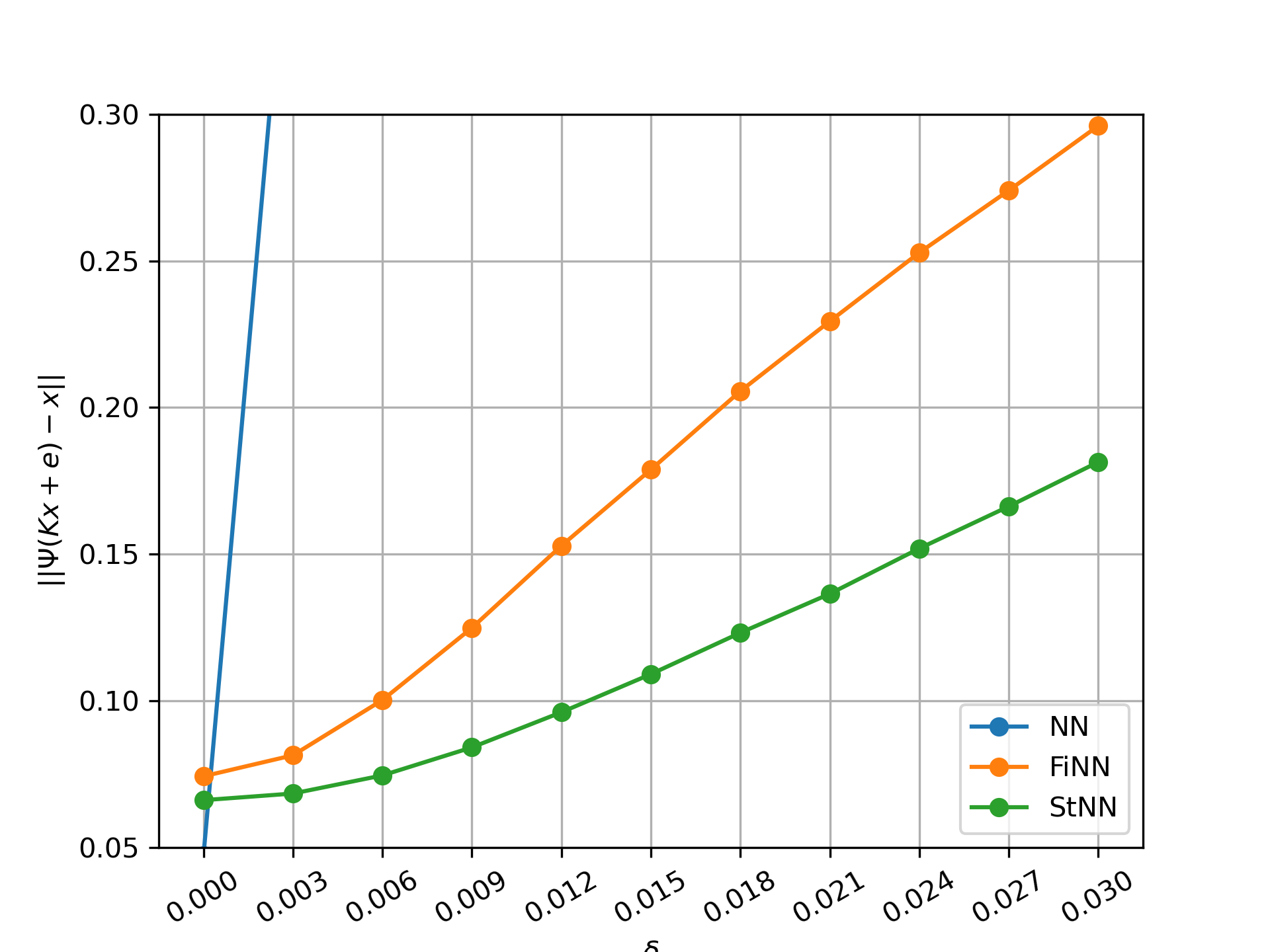}
& 
\includegraphics[width=0.3\textwidth]{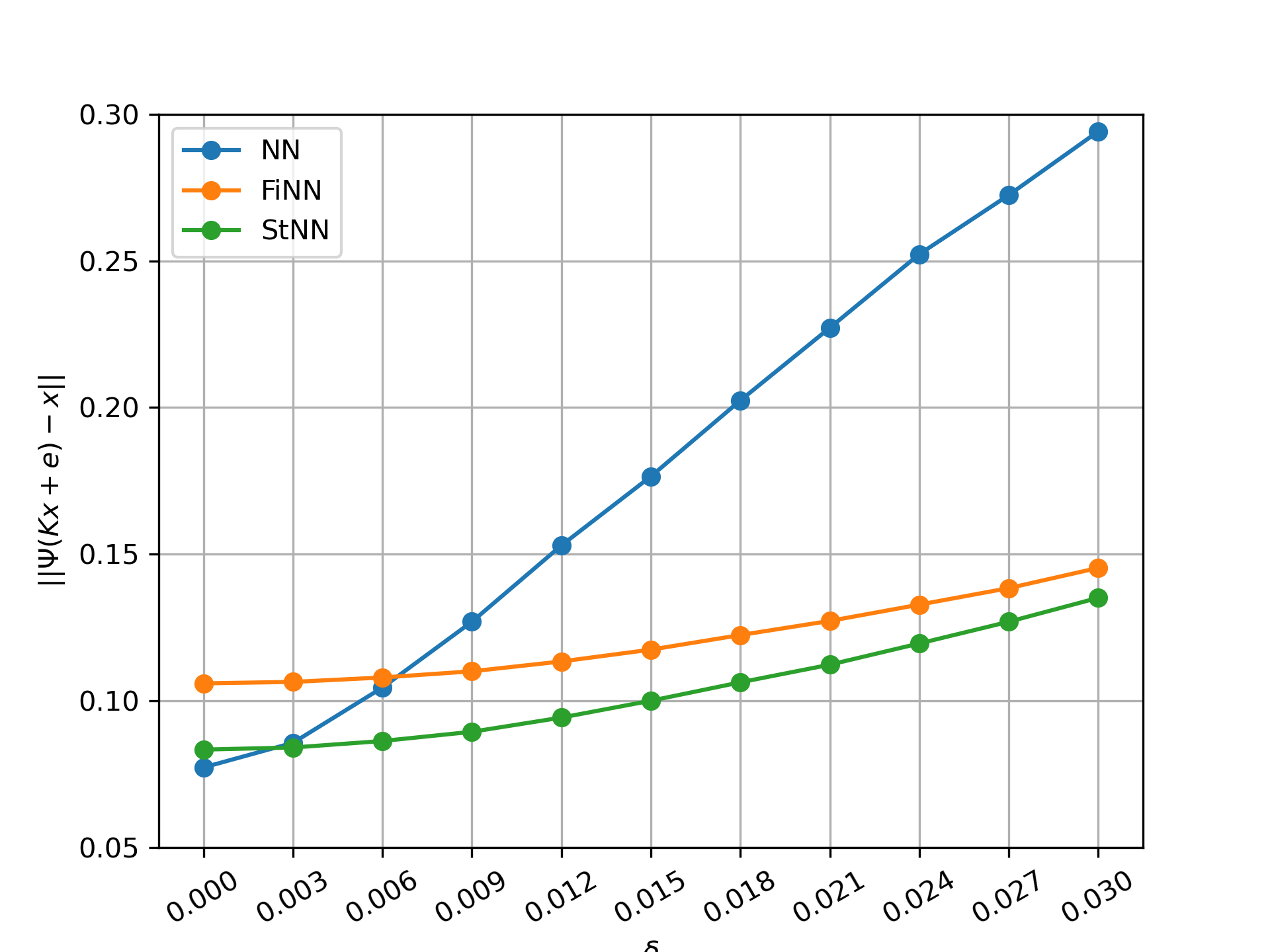} &
\includegraphics[width=0.3\textwidth]{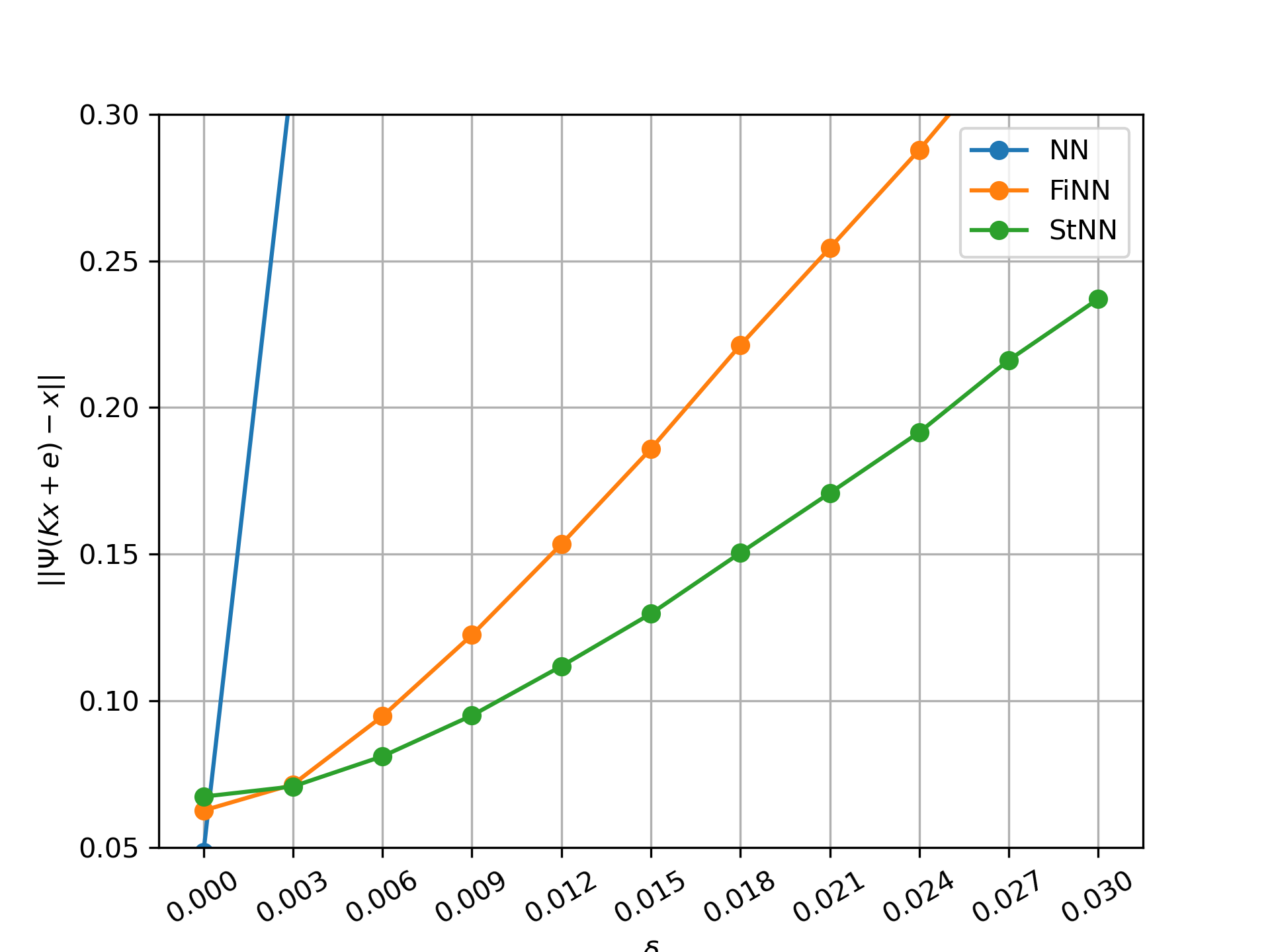} \\
\includegraphics[width=0.3\textwidth]{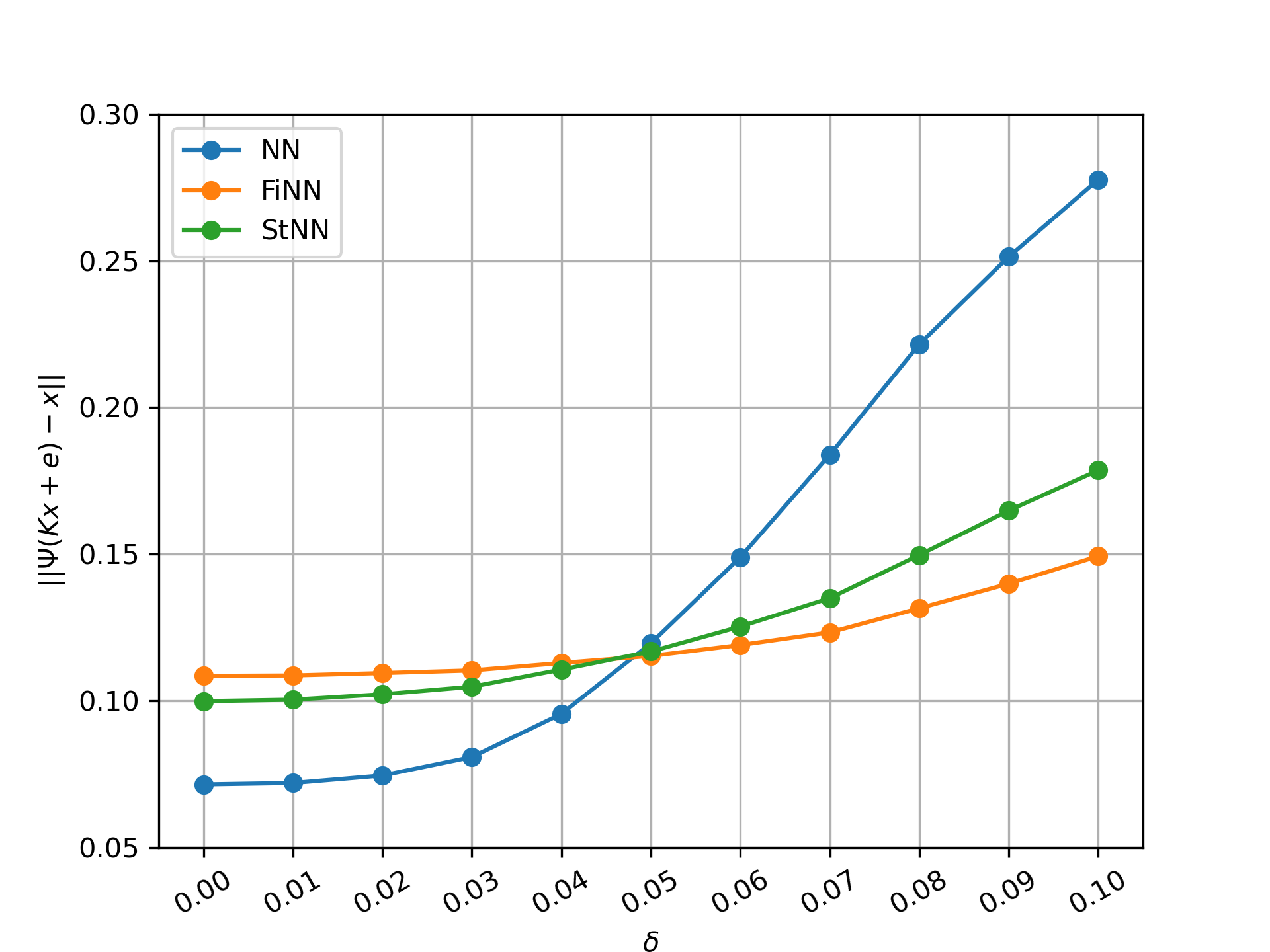}
& 
\includegraphics[width=0.3\textwidth]{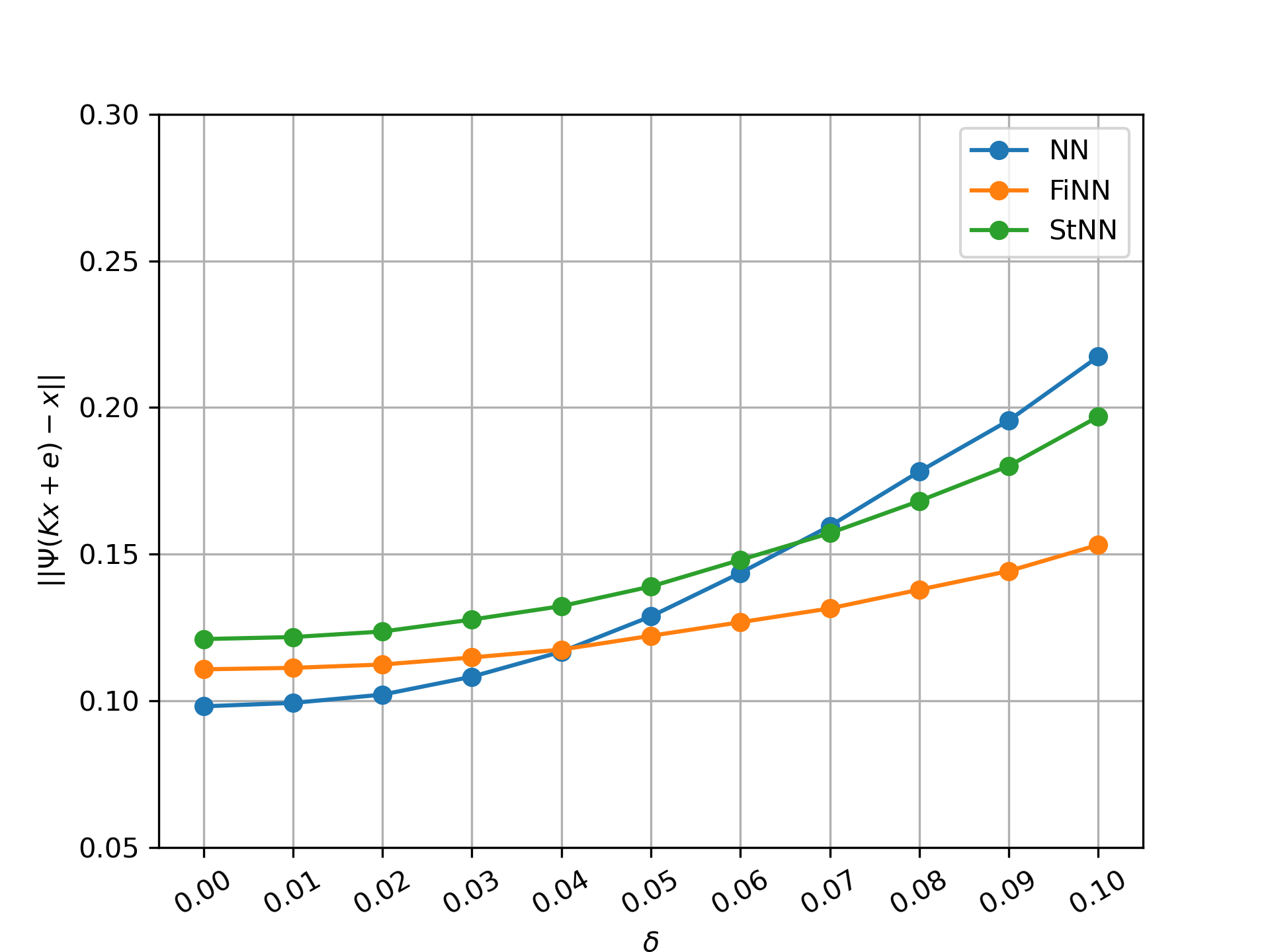} &
\includegraphics[width=0.3\textwidth]{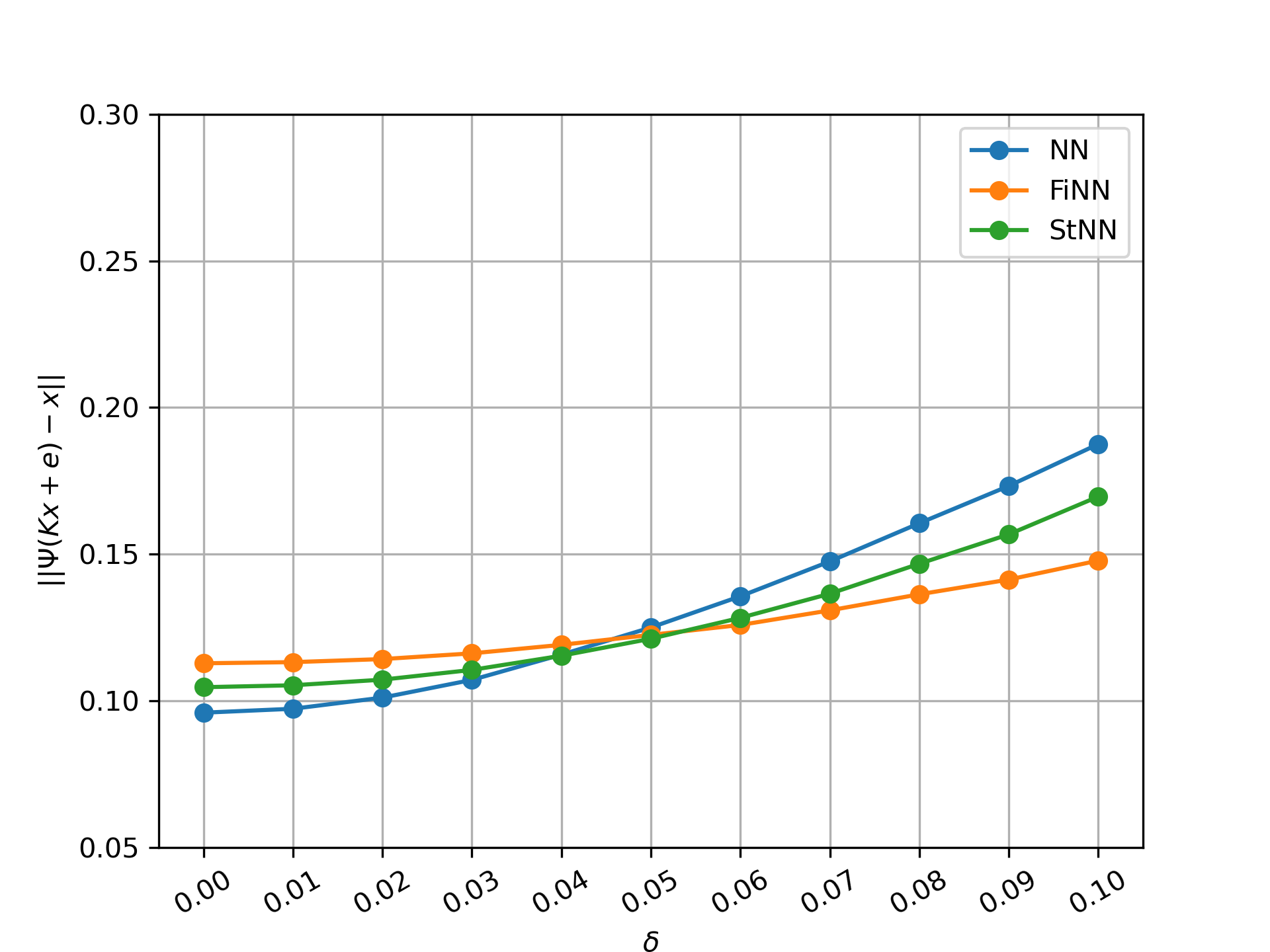} \\
\end{tabular}
\caption{Plots of the absolute error vs. the variance $\sigma$ of the noise for one image in the test set.
Upper row: experiment A. Lower row: experiment B.}
\label{fig:graficiNoise}
\end{figure}

In Figure \ref{fig:graficiNoise} we  plot, for one image in the test set, the absolute error between  the reconstruction and the true image vs. the noise standard deviation  $\sigma$. In the upper row the results from experiment A (we remark that in this experiment we trained the networks on no noisy data). The NN error (blue line) is out of range for very small values of $\sigma$ for both UNet and NAFNet, whereas the 3L-SSNet is far more stable. In all the cases, the orange and green line shows that  FiNN and StNN improve the reconstruction error. In particular, StNN performs best in all these tests.

Concerning experiment B (in the lower row of the figure), it is very interesting to notice that when the noise is smaller than the training one (corresponding to $\sigma=0.025$) the NN methods are the best performing  for all the considered architectures.  When $\sigma \simeq 0.05$ the behaviour changes and the stabilized methods are more accurate.


\section{Conclusions}\label{sec:conclusion}
Starting from the consideration that the most popular neural networks used for image deblurring, such as the family of convolutional UNets, are very accurate but unstable with respect to noise in the test images, we have proposed two different approaches to get stability without losing too much accuracy. The first one is a very light neural architecture, called 3L-SSNET, and the second one is to stabilize the deep learning framework by introducing a pre-processing step. 
Numerical results on the GoPro dataset  have demonstrated the efficiency and robustness of the proposed approaches, under several settings encompassing in-domain and out-of-domain testing scenarios. 
The 3L-SSNet overcome UNet and NAFNet in every test where the noise on test images exceeds  the noise on the training set, combining the desired characteristics of execution speed (in a green AI perspective) and high  stability.
The FiNN proposal increases the stability of the NN-based restoration (the values of its SSIM do not change remarkably in all the experiments), but the restored images appear too smooth and few small details are lost somewhere.
The StNN proposal, exploiting a model-based formulation of the underlying imaging process, achieves the highest SSIM values in the most challenging out-of-domain cases, confirming its great theory-grounded potential. It represents, indeed, a good compromise between stability and accuracy.
We finally remark that the proposed approach can be simply extended to other imaging applications modeled as an inverse problem, such as super-resolution, denoising, or tomography, where the neural networks learning the map from the input  to the ground truth image cannot efficiently handle noise in the input data. \\
This work represents one step further in shedding light on the black-box essence of NN-based image processing. 

\paragraph{Acknowledgments}
This work was partially supported by the US National Science Foundation, under grants DMS 2038118 and DMS 2208294.

\paragraph{Conflict of Interests}
The authors declare no conflict of interest.

\bibliographystyle{unsrt}
\bibliography{biblio}
\end{document}